\documentclass[11pt]{article}

\usepackage[final]{acl}

\usepackage{times}
\usepackage{latexsym}
\usepackage{amsmath}
\usepackage{url}
\usepackage[T1]{fontenc}

\usepackage[utf8]{inputenc}

\usepackage{microtype}

\usepackage{inconsolata}

\usepackage{graphicx}
\usepackage{subcaption}
\usepackage{linguex}
\usepackage{booktabs}
\usepackage{multirow}
\usepackage[table,xcdraw]{xcolor}

\title{When Context Misleads: Surprisal, Energy and Attention Entropy as Metrics of Coherence Illusions in LLMs}

\author{
 \textbf{Ece Takmaz},
 \textbf{Nitin Kumar},
 \textbf{Li Kloostra},
 \textbf{Jakub Dotlačil}
\\
 Utrecht University
\\
 \small{
   \textbf{Correspondence:} \href{mailto:e.k.takmaz@uu.nl}{e.k.takmaz@uu.nl}
 }
}

\begin{document}
\maketitle
\begin{abstract}
Psycholinguistic studies show that human readers fall for \textit{coherence illusions}: an incoherent discourse can seem coherent simply because a distractor matches what comes next. We investigate whether Dutch language models (6 monolingual and 4 multilingual) show the same behavior on texts that link back to earlier context with words such as `again' and `too'. First, we find that surprisal at the critical word tracks human acceptability judgments and eye-tracking data. Models are more surprised by incoherent continuations, but a matching distractor in the prior context reduces this surprisal. Second, attention entropy identifies heads that behave differently under coherence vs.\ incoherence. We find that ablating these heads shows transfer effects across experiments, suggesting a shared mechanism. Third, we introduce energy from the associative-memory literature as a metric to quantify discourse coherence. Taken together, our results show that coherence illusions arise in Dutch LLMs, with entropy and energy exposing mechanisms that operate across settings.

\end{abstract}

\section{Introduction}
Discourses often include words, called \textbf{presupposition triggers}, which increase the coherence of the text by tying what is being said to what has been said before \citep{sep-presupposition}. One such word is `again'. For instance, upon reading `\textit{John ate an apple as a morning snack and after lunch, he again had an apple}', `again' helps us link the event in the second sentence with what we have read so far. From the processing perspective, presupposition triggers force readers to retrieve chunks of information from memory. The retrieval from memory is not perfect, however. For instance, in `\textit{John ate an apple, Mary ate a pear; after lunch, John again had a pear}', the existence of the distractor of Mary eating a pear could cause interference due to the similarity in the items and events being mentioned, leading the reader into thinking that John might as well have eaten a pear again.

These failures in interpreting the discourse are called `\textbf{coherence illusions}', since  the reader comes under the illusion that the story so far has been coherent when it was not \citep{SCHMITZ2025104637}. These illusions are similar to  illusions of grammaticality~\citep{Phillips20115GI} such as subject-verb agreement attraction at the sentence level~\citep{BOCK199145}, which have been evidenced in a long line of research in psycholinguistics.

The sensitivity of language models to sentence-level illusions of grammaticality
 has been investigated in various types of models for a decade now~\citep{linzen-etal-2016-assessing},  also using benchmarks for acceptability in Dutch, the language we study in this paper~\citep{gronlp_2024, suijkerbuijk-etal-2025-blimp, jumelet-etal-2026-multiblimp}.

In this work, we go beyond sentence-level illusions and study whether LLMs predict \textbf{coherence illusions in short discourses in Dutch}. \citet{SCHMITZ2025104637} and \citet{li2026memory} investigate coherence illusions in Dutch and  show that these illusions are clearly visible in reading eye-tracking data, and, to a lesser extent, in acceptability judgements. In particular, when the discourse is incoherent, distractors matching the critical word cause interference. Intuitively, the readers appear to lose track of the details of actions, referents and subjects, and can come under the illusion of coherence, evidenced by facilitated reading in eye-tracking data and high acceptability judgments.

The presence of distractors in prompts has also been shown to cause confusions in LLMs, where a single random word can lead the LLM astray~\citep{elephants, fredashi, yoran2024making, niu-etal-2025-llama}. When a word is mentioned in the prompt, the model could assign higher probabilities to it later, even when that word might lead to factually incorrect responses. This behavior called \textbf{`contextual entrainment'} or \textbf{`context hijacking'} resembles coherence illusions. In this paper, we study the extent to which similar effects of coherence illusions in discourse emerge in LLMs using a range of methods to illustrate behavioral and model-internal findings. Our contributions are:
\begin{itemize}
    \item We find evidence for coherence illusions in LLMs that were exposed to Dutch (6 mono- and 4 multilingual models), on stimuli from 3 psycholinguistics experiments.
    \item We investigate  surprisal, attention entropy, and introduce `energy' to quantify discourse coherence in LLMs. 
    \item We find that surprisals at the critical position are in line with human reading eye-tracking data and acceptability judgements. 
    \item Attention entropy helps  identify heads that are informative across coherence conditions.
    \item Using linear probes, we find that models encode information about discourse coherence.
\end{itemize}

Language models have long been studied from the perspective of theories of human language processing and to what extent they can parallel findings from psycholinguistics. This link forms the basis of our motivation to study a phenomenon that was evidenced in human studies, coherence illusions, and a phenomenon evidenced in LLMs, contextual entrainment, allowing us to investigate the plausibility of LMs as models of human language processing. Furthermore,  combined with the findings from interpretability studies of LLMs and how context might cause models’ output to go in the wrong direction, our findings also inform us about failure cases of LMs and give us pointers about the mechanisms underlying the computations leading to erroneous behavior in LMs. 

Coherence illusions in humans and contextual entrainment in LLMs have been studied in isolation, yet both can be framed in terms of cue-based retrieval and similarity-driven interference~\citep{cuebased}. This unifying lens provides a conceptual link to connect research lines that have so far been treated separately.\footnote{Code available at \url{https://github.com/ecekt/coherence_illusions_llms}.}

\section{Related Work}

\subsection{Coherence Illusion in Humans}
Discourse-level illusions in humans have recently been studied by \citet{SCHMITZ2025104637} and \citet{li2026memory} in Dutch. Their experiments utilize \textbf{presupposition triggers} \textit{ook} `too' and \textit{opnieuw} `again' in short discourses that prompt the readers to connect upcoming information with prior context. The authors study the effects of distractors, negative scope, and coherence within the discourse. \citet{SCHMITZ2025104637} find  that a semantically matching but inaccessible (due to negation) distractor can still facilitate processing as evidenced by eye-tracking reading data. \citet{li2026memory} show that, in incoherent discourses, distractors matching the critical word  facilitate reading,  and, in one experiment, increase acceptability. Both studies find evidence for \textbf{coherence illusions} in Dutch.  These findings are explained by \textbf{cue-based retrieval theory} ~\cite{cuebased}, where cues activate items in memory based on  similarity. Even partially-matching, negated distractors cause facilitatory interference, as evidenced in eye-tracking data and in some acceptability judgment experiments.

\subsection{LMs as Tools for Psycholinguistics and as Cognitive Models}

\citet{ryu-lewis-2021-accounting} and \citet{RYU2025104670} draw parallels between  the attention mechanism in transformers~\cite{vaswani} and cue-based retrieval accounts of language processing \cite{cuebased, VANDYKE2003285}. The queries are cues, values are items stored in memory. The entropy of the attention distribution, \textbf{`attention entropy'}~\citep{ryu-lewis-2021-accounting, RYU2025104670}, has been linked to  linguistic phenomena such as illusions of grammaticality (agreement attraction) and center-embedding asymmetries~\citep{oh-schuler-2022-entropy}. The diffuseness of attention can be a sign of interference, where candidates from the context compete to be retrieved. 

An alternative account of language processing is the expectation-based `\textbf{surprisal theory}'~\citep{hale2001, LEVY20081126}, which attributes difficulty to a word's predictability as quantified by information-theoretic surprisal~\cite{shannon}, which is the negative log probability of a word $w_i$ given its previous context: $\text{surprisal}(w_i) = - \log_2 P(w_i | w_{<i})$. Surprisal has been the focus of many studies exploring the relation between LM-derived metrics and  behavioral data from psycholinguistics~\citep{wilcox2023testing, shain, tsipidi2026probingreadingtimes}.

However, LMs also exhibit notable misalignments with human processing: \textit{larger} language models trained on much more data fit reading signals worse than smaller predecessors trained on smaller datasets \cite{byungdoh, kuribayashi-etal-2024-psychometric}, with some variation across the layers~\cite{internal}. Related to this, \citet{von2026diverging} show mixed results for the match between transformers and human data on illusions of grammaticality.%

\subsection{Context Effects in LLMs}
\citet{niu-etal-2025-llama} show that LLMs assign tokens present in the prompt significantly higher probabilities regardless of their relevance even if this produces factually incorrect continuations. They call this phenomenon `\textbf{contextual entrainment}' and identify heads contributing to this behavior, which is along the lines of earlier work where \citet{fredashi} and \citet{yoran2024making} had shown that LLMs can be negatively influenced by irrelevant contexts, with errors accumulating over time. \citet{elephants} call this type of behavior `\textbf{context hijacking}', demonstrating how easily fact-retrieval can be manipulated by prepending words to prompts in LLMs. \citet{elephants} attribute context hijacking in LLMs to \textit{associative memory retrieval}, where tokens in a prompt act as cues and retrieval can be manipulated by distractors. 

\subsection{Associative Memory and Energy}
\label{relenergy}
Transformer attention has been linked to  \textbf{associative memory} from a machine learning perspective in addition to human language processing, which motivates our study. This view connects transformer attention to classical Hopfield Networks~\citep{hopfield1982neural, hopfield1984neurons}, dense associative memory~\citep{krotov2016dense} and modern continuous Hopfield networks~\citep{ramsauer2021hopfield}. The idea behind memory retrieval in these models is traversing an `\textbf{energy}' landscape. Given an input query, the network tries to retrieve an item from memory most fitting the input. Some recent works propose architectural changes to accommodate energy and associative memory structures in LLMs explicitly~\citep{nrgpt, zanzotto-etal-2025-position}. The energy measure proposed by \citet{nrgpt} is shown to relate to both attention entropy and surprisal in predicting reading difficulty~\citep{dotlacil2026energybasedtransformerspredictorsreading}.

\subsection{Interpretability} 
The research on interpretability offers various techniques to find components that contribute to linguistic processes in LMs. For instance, \citet{boguraev-etal-2025-causal} finds that filler-gap constructions share underlying mechanisms in LMs. Similarly, \citet{wang2023interpretability} investigates indirect object identification using minimal pairs to find how certain heads contribute to locating duplicates and copying behavior, connecting to 
induction heads~\citep{elhage2021mathematical, olsson2022context}. %

\section{Methodology}
To test whether LMs and humans display similar behavior when processing coherent vs.\ incoherent texts, we use existing sets of sentences created for psycholinguistics experiments that collected human reading signals focusing on coherence illusions in Dutch~\citep{SCHMITZ2025104637, li2026memory}. In Section~\ref{data}, we go into detail about the sentence types, and in Section~\ref{models}, we list the models and the metrics we derive from them.

\begin{table*}[]
\small
\centering\begin{tabular}{@{}lll@{}}
\toprule
\multicolumn{3}{c}{Megan had honger maar gelukkig stond er een volle fruitschaal in de lobby van het hotel.}\\
\multicolumn{3}{c}{Megan was hungry but luckily there was a full fruit bowl in the hotel lobby.}\\ \midrule
\rowcolor[HTML]{FFFFFF} 
{\color[HTML]{27272A} \textbf{Condition}} & \multicolumn{1}{c}{\cellcolor[HTML]{FFFFFF}{\color[HTML]{27272A} \textbf{Target}}} & \multicolumn{1}{c}{\cellcolor[HTML]{FFFFFF}{\color[HTML]{27272A} \textbf{Distractor}}} \\ \midrule
\textbf{MAMA}                             & \multicolumn{1}{c}{Megan heeft een gele \textbf{appel} gegeten,}                            & \multicolumn{1}{c}{maar ze heeft geen groene \textbf{appel} gegeten.}                           \\
& \multicolumn{1}{c}{Megan ate a yellow apple,} & \multicolumn{1}{c}{but she did not eat a green apple.}\\ \midrule
\textbf{MAMI}                             & \multicolumn{1}{c}{Megan heeft een gele \textbf{appel} gegeten,}                                                & \multicolumn{1}{c}{maar ze heeft geen groene \textbf{peer} gegeten.}            \\
& \multicolumn{1}{c}{Megan ate a yellow apple,} & \multicolumn{1}{c}{but she did not eat a green pear.}                                     \\ \midrule
\textbf{MIMA}                             & \multicolumn{1}{c}{Megan heeft een gele \textbf{peer} gegeten,}                                                 & \multicolumn{1}{c}{maar ze heeft geen groene \textbf{appel} gegeten.}   \\
& \multicolumn{1}{c}{Megan ate a yellow pear,} & \multicolumn{1}{c}{but she did not eat a green apple.}                                             \\ \midrule
\textbf{MIMI}                             & \multicolumn{1}{c}{Megan heeft een gele \textbf{peer} gegeten,}                                                 & \multicolumn{1}{c}{maar ze heeft geen groene \textbf{peer} gegeten.}   \\
& \multicolumn{1}{c}{Megan ate a yellow pear,} & \multicolumn{1}{c}{but she did not eat a green pear.}                                              \\ \midrule
\multicolumn{3}{c}{De ochtend erna heeft Megan \textbf{opnieuw} een \textbf{appel} gegeten want ze wilde meer fruit eten.}\\
\multicolumn{3}{c}{Next morning, Megan \textbf{again} ate an \textbf{apple} because she wanted to eat more fruit.}\\
\bottomrule
\end{tabular}
\caption{Sample sentences from the `Again' study 1. The experiment had an introductory sentence (top), followed by one of the  conditions, which was followed by the critical sentence in which the presupposition trigger appears. Note that, in Dutch, `again' is immediately followed by `an apple', and the negation in these sentences comes from the word `geen'.}
\label{tab:nlsents}
\end{table*}

\subsection{Data}
\label{data}
The data originates from 5 experiments conducted with Dutch texts, 3 of which we use in this paper~\cite{SCHMITZ2025104637, li2026memory}. These texts narrate a short story, usually in 3 sentences. The first sentence establishes the context. The second sentence contains two propositions, introducing the \textbf{target} and the \textbf{distractor}. The final sentence includes either `again' or `too' as a trigger. The \textbf{critical} word following the presupposition trigger makes the discourse either coherent or incoherent depending on the target. We explain the 2 studies that use `again'. The `Too' studies were created in a parallel way to the `Again' Study 1, see Appendix~\ref{app_too} for samples.

\subsubsection{`Again' Study 1}
\label{again}
The discourse narrates a story about a single subject and involves the presupposition trigger  `again' (`opnieuw' in Dutch) in 36 stimulus sets each containing 4 discourses, in total, 144 samples.\footnote{Although the number of sentences seems low for researching LMs, similar studies make use of a limited number of sentences (e.g., \citet{arora-etal-2024-causalgym} use 50 sentences; \citet{boguraev-etal-2025-causal} use 400 sentences to evaluate a 1.4B model, and 96 sentences for  2.8B and 6.9B models).} 
We use the following notation: \textbf{MA} represents a `match' with the critical word and \textbf{MI} represents a `mismatch' with the critical word. The first half indicates the \textit{target-critical} relation and the second one indicates the \textit{distractor-critical} relation. As an example, \textbf{MIMA} represents the coherence illusion case, where the target mismatches the critical word, but the distractor matches it. See Table~\ref{tab:nlsents} for an example stimulus set. The stimuli consist of samples with a wide variety of topics to test a general tendency in humans. See Appendix~\ref{app_too}, for a sample that differs in semantics. The actions, subjects and objects differ substantially (e.g., buying, seeing, drawing, interviewing, receiving, practicing and so on). We used the apples and pears example throughout the paper for consistency and ease of reading.

\subsubsection{`Again' Study 2}
\label{again2}
This study is similar to the `Again' Study 1, but introduces 2 subjects in the discourse to manipulate target/distractors. An example of the MIMA condition is given here, only in English:
\begin{itemize}
    \item \textbf{MIMA} The \textbf{manager} and the \textbf{assistant} got a piece of fruit from the canteen during the break. The \textbf{manager} ate a yellow \textit{pear}, and the \textbf{assistant} ate a green \textit{apple}. The \textbf{manager} again ate an \textit{apple} later that afternoon because she wanted to eat more fruit.
\end{itemize} 

In addition, to study distance effects, the order of the target and distractor was varied: either the target preceded the distractor, as above, or their order was reversed. The first condition was coded as RMIMA, the second was coded as PMIMA. The rest of the conditions were constructed in the same way as in the `Again' study 1. The data consists of 36 sets of 8 discourses, 288 in total. See Table~\ref{fig:dutch_discourse_2referents} for samples in Dutch.

\subsection{Models}
\label{models}

We use a set of open-source models that included Dutch data in their training or fine-tuning. As relatively small Dutch-only language models, we use models based on the GPT-2~\citep{radford2019language} and GPT-neo~\citep{gpt-neo} architectures.\footnote{
GPT-2-based models are still relevant and valid in psycholinguistics, since the surprisal extracted from them has been found to better predict human reading times compared to larger LMs trained on larger corpora~\citep{byungdoh}.}$^,$\footnote{HuggingFace names: `yhavinga/gpt2-medium-dutch', `yhavinga/gpt2-large-dutch', `yhavinga/gpt-neo-125M-dutch', `yhavinga/gpt-neo-1.3B-dutch',  `GroNLP/gpt2-small-dutch'.} We employ the set of models by \textbf{yhavinga} (based on GPT-2 and GPT-neo), with the creator's recommended model being `yhavinga/gpt2-medium-dutch'. From the \textbf{GroNLP} family of models, we utilize the recommended `GroNLP/gpt2-small-dutch' model~\citep{de-vries-nissim-2021-good}. As an LLM trained on Dutch text, we use the \textbf{GEITje} model `BramVanroy/GEITje-7B-ultra', which was based on Mistral~\citep{vanroy2023language, vanroy2024geitje7bultraconversational}.

We then select a set of multilingual LLMs. We use the \textbf{EuroLLM} model `utter-project/EuroLLM-9B-2512' \citep{martins2025eurollm9btechnicalreport, MARTINS202553}. This LLM supports 24 official languages of the European Union, including Dutch. We also use the \textbf{Aya Expanse} model that explicitly lists Dutch in its coverage list `CohereLabs/aya-expanse-8b'~\citep{dang2024ayaexpansecombiningresearch}. Additionally, we use the strong multilingual model from the \textbf{Qwen} family `Qwen/Qwen3-8B' that covers Dutch~\citep{yang2025qwen3technicalreport}. Finally, we use `meta-llama/Llama-3.1-8B' from the \textbf{LLaMA} family, although Dutch is not listed~\citep{grattafiori2024llama3herdmodels}. See Appendix~\ref{expsetup} for the processing of texts and the experimental setup.

\section{Surprisal of the Critical Word}
\label{surprisal}
As the first experiment, we focus on the next-word prediction task. We extract surprisal values of the critical word, right after seeing the presupposition triggers. For the `Again' Studies, we obtain the surprisal of the  critical word after `again a/an'. In Dutch, it would be the surprisal of the observed word, e.g., `appel' after seeing `opnieuw \textit{een}'. For the `Too' Study, we extract the surprisal of the observed critical word  after seeing `\textit{ook}'. In case the critical word is tokenized into multiple tokens, we look at the first token's surprisal. We plot the mean surprisals per model per condition.

\subsection{Surprisal in `Again' Studies}
\label{surpagain}

\begin{figure}[h]
    \centering
    \includegraphics[width=0.9\linewidth]{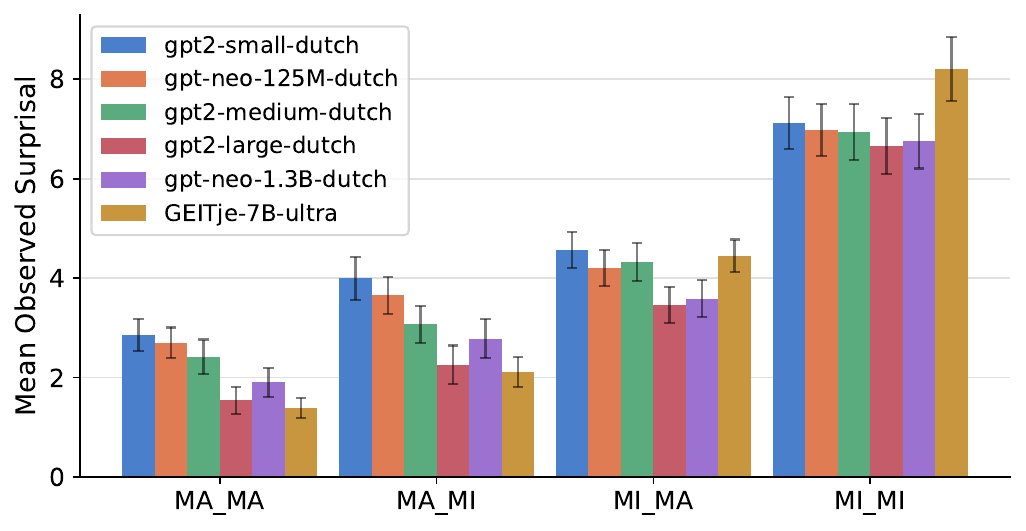}
  \caption{Surprisals for the `Again' Study 1 for the Dutch-only models. Error bars show the Standard Error.}
  \label{fig:supexp1opnieuwdutch}
\end{figure}

\begin{figure}[h]
    \centering
    \includegraphics[width=0.9\linewidth]{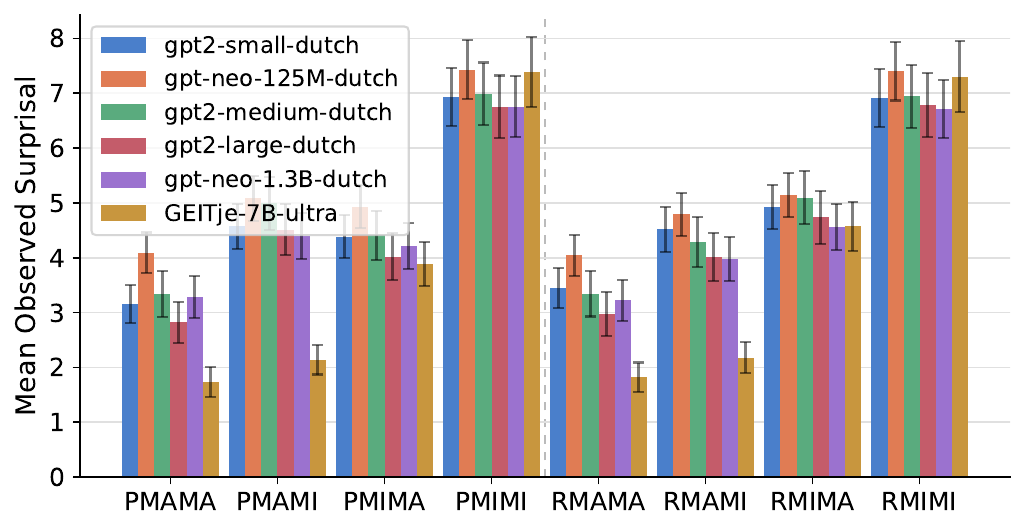}
  \caption{Surprisal results for the `Again' Study 2 for the Dutch-only models.}
  \label{fig:supexp2opnieuwdutch}
\end{figure}

\noindent\textbf{Dutch-only models.} Figures \ref{fig:supexp1opnieuwdutch} and \ref{fig:supexp2opnieuwdutch} depict the surprisals for `Again' Study 1 and 2, respectively. We see that the Dutch-only models show a pattern where surprisals increase in this order: MAMA < MAMI < MIMA < MIMI. MIMA is the coherence illusion case, and it shows smaller surprisals compared to MIMI, although it is also incoherent. The results of related samples t-tests reveal significant differences across all conditions in all models, $p<.001$. Therefore, the results are aligned with our expectations based on empirical findings from psycholinguistics studies. This also agrees with the contextual entrainment findings, where the existence of a distractor token in the input prompt makes its likelihood higher in later positions \citep{niu-etal-2025-llama}. Similarly, for the `Again' Study 2, PMIMI and RMIMI have the highest surprisal, and MIMA versions differ significantly from them. The ordering of the discourse does not seem to have a significant effect (P vs.\ R) on the surprisals in the Dutch-only models, like the findings in human studies~\citep{li2026memory}. Larger GPT models show less surprisal in general. GEITje, the Dutch-only LLM, yields more surprisal in incoherent discourses compared to the other Dutch-only models. 

\begin{figure}[h]
    \centering
    \includegraphics[width=0.9\linewidth]{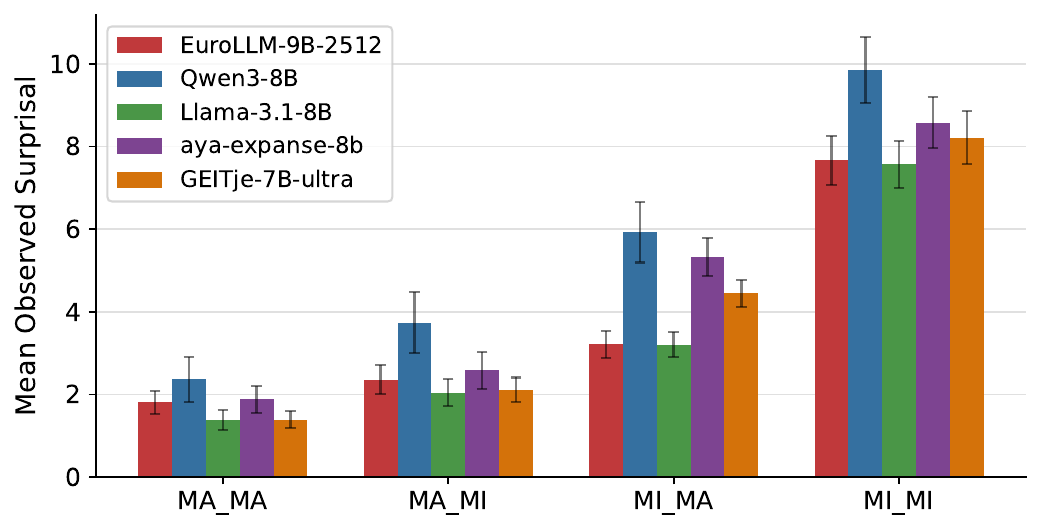}
  \caption{Surprisal results for the `Again' Study 1 for the LLMs.}
  \label{fig:supexp1opnieuwllms}
\end{figure}

\noindent\textbf{Multilingual LLMs.} Figure \ref{fig:supexp1opnieuwllms} depicts surprisals for `Again' Study 1. We see that the pattern from the Dutch-only models is repeated also in LLMs, where surprisals increase in this order: MAMA < MAMI < MIMA < MIMI, with significant differences across conditions, $p<.001$. The coherence illusion case of MIMA is also visible here, aligning with our expectations based on empirical findings from psycholinguistics studies. The results for `Again' Study 2 with LLMs agree with those of Dutch-only models, see Figure~\ref{fig:supexp2opnieuwllms} in Appendix. We note that Qwen3 and, to a smaller extent, Aya Expanse exhibit more prominent surprisal increase in incoherent discourses. 

\noindent\textbf{Comparison to eye-tracking data.} Since empirical data exists from human reading studies~\citep{li2026memory}, we also compare surprisal against human gaze in `Again' Study 1 and 2. We use GPT-Neo-125M and GEITje as Dutch-only models for this comparison. We parse the final sentence into four regions, see Table~\ref{fig:four_regions_for_analysis}. The first region contains the presupposition trigger, then, the Critical region and the Post-critical region (Spillover). %

We then fit separate Linear Mixed Effect Models (LMEMs) for 3 eye-fixation duration measures, which are commonly utilized in psycholinguistic studies and when comparing LLM behavior to that of humans: first fixation duration (duration of the very first fixation on a word, FF), total gaze duration (sum of the durations of fixations that fall within a region before the region is left progressively for the first time, TG), and total fixation duration (sum of all fixation durations on a region, TOTFIX). %
We fit  a model with z-normalized surprisal as a continuous predictor and subjects and items as random effects. We start  with a maximal model and simplified it until convergence was achieved. We also fit  a model with surprisal, Target and Distractor as fixed effects. Both models are presented in Appendix~\ref{app:lmem-more}.

For GEITje in Study 1, surprisal is a significant predictor for all three fixation measures. This effect is significant even for the first fixation-duration measure, suggesting that surprisal captures early-stage difficulty rather than rereading measures alone. Thus, we see that surprisal is a reliable proxy for human reading difficulty during presupposition resolution, consistent with \citet{wilcox2023testing}, but now demonstrated at the discourse level. 
For GEITje in Study 2, surprisal effects are significant across all three measures in the critical region, with larger coefficients than in Study 1. The results for the GPT-Neo-125M  model are very similar to those of  GEITje, with surprisal in the critical region serving as a reliable proxy for human reading difficulty at the discourse level.

We also computed the surprisal for each token in each region for the two models. We report total and mean surprisal, along with standard errors, in Table \ref{tab:surprisal_by_region}. Analysis of surprisal in the object region showed that matching distractors caused a decrease in reading times overall, regardless of whether the discourse was coherent or incoherent. In the incoherent discourse, the surprisal for matching distractors was lower than for mismatching distractors, indicating a facilitatory effect which supports the coherence illusion. We note this consistently across both experiments for both models. %

\subsection{Surprisal in the `Too' Study}
\label{surptoo}
\begin{figure}[h]
    \centering
    \includegraphics[width=0.9\linewidth]{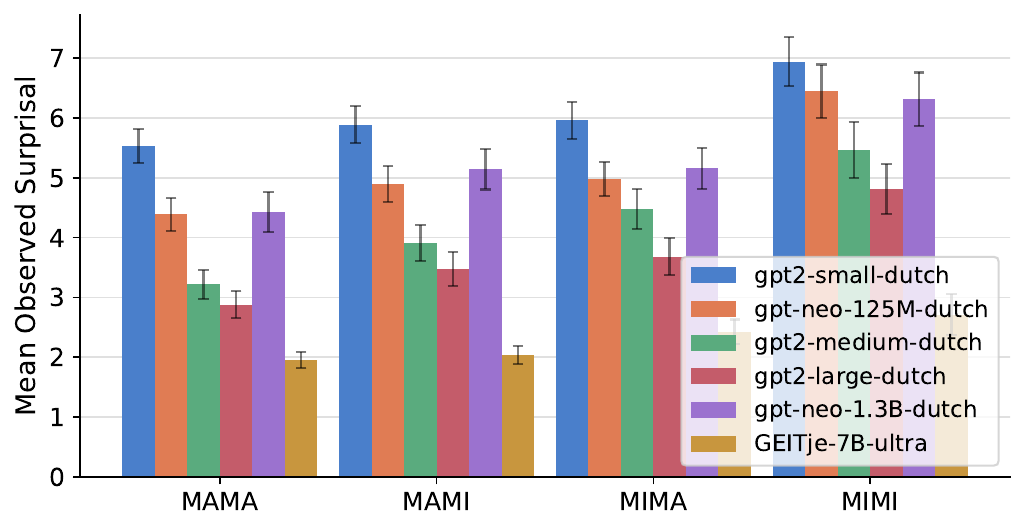}
  \caption{Surprisal results for the `Too' Study for the Dutch-only models.}
  \label{fig:supexp1toodutch}
\end{figure}

\noindent\textbf{Dutch-only models.} Figure  \ref{fig:supexp1toodutch} depicts the surprisals for `Too' Study. We see that the Dutch-only models show a surprisal pattern of MAMA < MAMI < MIMA < MIMI. The differences are mainly significant across conditions, with the exception of MAMI and MIMA, for which the models yield similar surprisals. 

\noindent\textbf{Multilingual LLMs.} LLMs show a larger surprisal increase towards MIMI, suggesting that they are more sensitive to incoherence, see Figure~\ref{fig:supexp1toollms}. The outcome of MIMA receiving less surprisal than MIMI suggests that coherence illusions are present in this case.

\section{Attention Entropy} 
\label{attent}
Attention distributions can differ across conditions, as the model aggregates the previous tokens into a contextualized representation. We are thus interested in the \textbf{entropy of the attention distribution}. 
To analyze the diffuseness of the words retrieved from the discourse and utilized before predicting the critical word, we extract the attention applied by the  trigger. We investigate attention entropy per head per layer across coherence conditions.

We calculate each head's ($h$) attention weights to the words in the context at the trigger position ($t$) per layer ($l$) (discarding the attention weight to self and normalizing the rest). Afterwards, we compute Shannon entropy, $\text{AttnEntr}_{l,h}(w_t)$:
\begin{equation}
 -\sum_{j=1}^{t-1} \text{Attn}_{l,h}(w_t, w_j) \times \log_2 \text{Attn}_{l,h}(w_t, w_j)    
\end{equation}

We choose the GPT-2 medium and large models as the Dutch-only models and the EuroLLM model as the multilingual model to investigate attention entropy. We extract attention entropies for `Again' Studies 1\&2 and the `Too' Study. Here, we focus on the coherence illusion case in the `Again' Study 1, MIMA, and we compare it to MIMI. Both of these conditions have target-critical mismatches; however, in MIMA, due to the matching distractor, human participants exhibit behavior reflecting perceived coherence. Given the surprisal results in Section~\ref{surpagain}, the models we tested also show much lower surprisal for MIMA than for MIMI, confirming coherence illusions in models. 

\begin{figure}[h]
    \centering
    \includegraphics[width=0.9\linewidth]{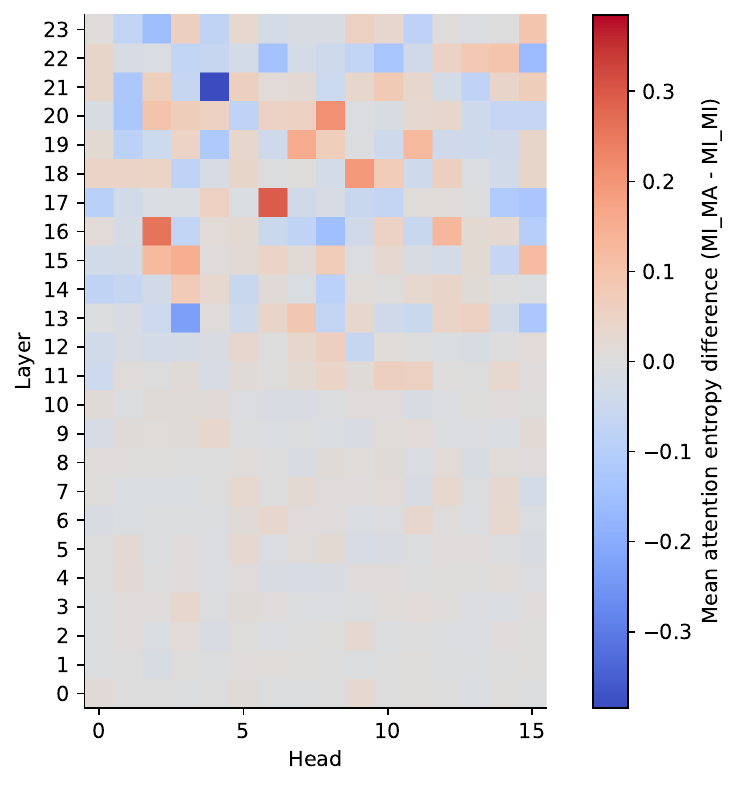}
  \caption{Attention Entropy difference (MIMA-MIMI) in `Again' Study 1 in the medium Dutch GPT-2 model's heads across layers. Positive values indicate higher entropy in the MIMA condition compared to MIMI.}
  \label{fig:attentrdiffmedium}
\end{figure}

Figure~\ref{fig:attentrdiffmedium} shows the difference in attention entropies between MIMI and MIMA in GPT-2-medium. Most of the heads that differ in behavior are located in the \textbf{latter} half of the layers. Some heads seem to suppress information related to coherence and others boost that information. It must also be noted that diffuse attention does not necessarily mean confusion or interference, it could also be needed to gather enough information  from the context  while processing the current token~\citep{RYU2025104670}. 

We provide additional entropy difference matrices comparing incoherent condition and the coherence illusion condition in Appendix~\ref{app_ent}. These matrices show that `Again' Study 2 yields a matrix similar to `Again' Study 1, `Too' Study 1 reveals smaller differences between conditions, and the large GPT-2 model has heads that differ in entropy across a larger proportion of the layers. EuroLLM model has the most distinct matrix, with even earlier layers showing differences, and the strongest differences are in the middle layers. The general trend seems to be that middle-to-late layers are activated in different coherence conditions, potentially indicating the role of increased contextualization in encoding coherence. 

\section{Energy as a Metric to Evaluate Coherence} 
We propose a metric that was not used in the psycholinguistic analysis of language models before: `\textbf{energy}' due to its connections to memory retrieval accounts, see Section~\ref{relenergy}. We use energy to quantify model behavior when faced with coherent vs.\ incoherent narratives. This connects to the cue-based retrieval literature in psycholinguistics by casting  presupposition resolution as the retrieval of relevant items from the preceding context. We use the energy function that was utilized by \citet{ramsauer2021hopfield} who showed connections between the transformer attention and memory retrieval in Hopfield Networks. $\mathbf{X}$ represents the keys in the memory and $\boldsymbol{\xi}$ is the query. $lse$ is the log-sum-exp function. $\beta$ is the inverse square root of the dimensions of the query. $N$ is the number of keys, $M$ is the maximum norm observed in the keys.
\begin{equation}
E = -\text{lse}(\beta, \mathbf{X}^T \boldsymbol{\xi}) 
  + \frac{1}{2} \boldsymbol{\xi}^T \boldsymbol{\xi} 
  + \beta^{-1} \log N 
  + \frac{1}{2} M^2
\end{equation}
\begin{equation}
    \text{lse}(\beta, \mathbf{X}^T \boldsymbol{\xi}) 
= \beta^{-1} \log \sum_{i=1}^{N} \exp(\beta \, \mathbf{x}_i^T \boldsymbol{\xi})
\end{equation}

We operationalize the energy calculation in two ways: (1) \textbf{Energy 1:} Query is the last token of the presupposition trigger, `again \textit{a/an}', (2) \textbf{Energy 2:} Query is the first token of the critical word, e.g., `apple' following `again an'. The keys in both cases originate from the tokens in the previous context. In this way, we aim to see how the energy would change based on the similarities between the queries and the keys in the context as well as the attention paid to them. If there is a very good match that the query attends to, the discourse will have a low energy score. If the query attends to multiple tokens with diffuse attention, the energy will be high.

We use the Dutch GPT-2 models (medium and large) and extract the queries and keys of the words in the discourse for the `Again' Study 1. We register forward hooks to extract query and key projections of hidden states right before they are fed into the attention computation. We take MIMI as the baseline condition as it is a mismatching condition with incoherence, and see how the energies in other conditions fare. We calculate the average energy per layer across all stimuli, and deduct MIMI's average per layer from these to offset the other conditions given this baseline. 

\begin{figure}[h]
    \centering
    \includegraphics[width=0.9\linewidth]{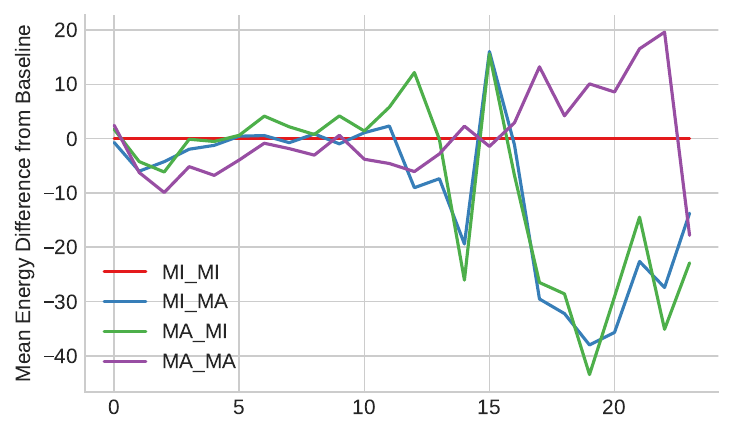}
  \caption{\textbf{Energy 1} scores across layers for the \textit{medium} Dutch GPT-2  model for the `Again' Study 1.}
  \label{fig:energy1exp1medium}
\end{figure}

Figure~\ref{fig:energy1exp1medium} depicts the \textbf{Energy 1 }scores for the medium GPT-2 model for the `Again' Study 1, see Figure~\ref{fig:energy1exp1large} in Appendix for the large model. We see that MAMA has higher energy especially in the later layers, whereas both MIMA and MAMI have much lower energies both in the medium and large models. The lower energy of the MAMI and MIMA conditions likely signals that the model can differentiate targets and distractors in those cases, which helps it with integrating the presupposition trigger, which is currently processed. In human studies,  participants go back and check \textit{specific} words in MIMA and MAMI, which might be aligned with less diffuse attention to words and low energy.

\begin{figure}[h]
    \centering
    \includegraphics[width=0.9\linewidth]{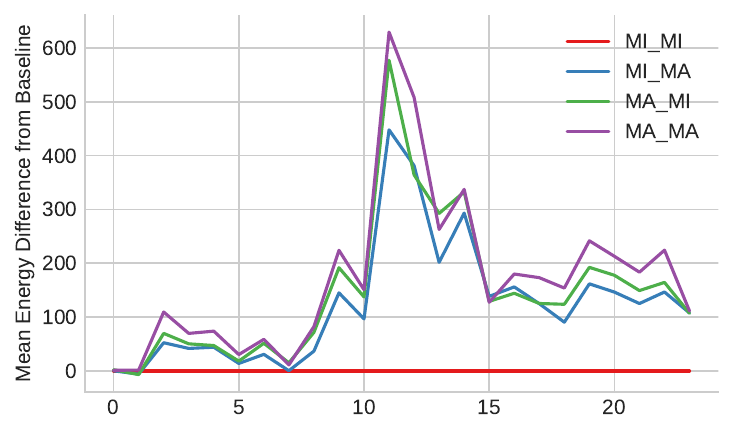}
  \caption{\textbf{Energy 2} scores across layers for the \textit{medium} Dutch GPT-2  model for the `Again' Study 1.}
  \label{fig:energy2exp1medium}
\end{figure}

Figure~\ref{fig:energy2exp1medium} depicts the \textbf{Energy 2} scores for the medium GPT-2 model, see Figure~\ref{fig:energy2exp1large} in Appendix for the large model. This score takes into account seeing the actual continuation to the presupposition trigger. We observe that MAMA, MIMA and MAMI have generally higher energies compared to MIMI, with MAMA having the highest energy across layers (as in Energy 1). The high energies observed in MAMA can be due to the fact that the target, distractor and critical word are the same words, causing the energy function to match multiple keys, therefore, leading to higher scores. 

In both energy types, we see diverging behavior between MIMI and MIMA, particularly in middle and later layers, which points to the existence of processing differences even though both conditions represent incoherent discourses. As the energy is useful in differentiating between conditions that vary in coherence and presupposition matches, we propose further investigations of this metric.

\section{Coherence Head Ablation}
\begin{figure}[h]
    \centering
    \includegraphics[width=0.9\linewidth]{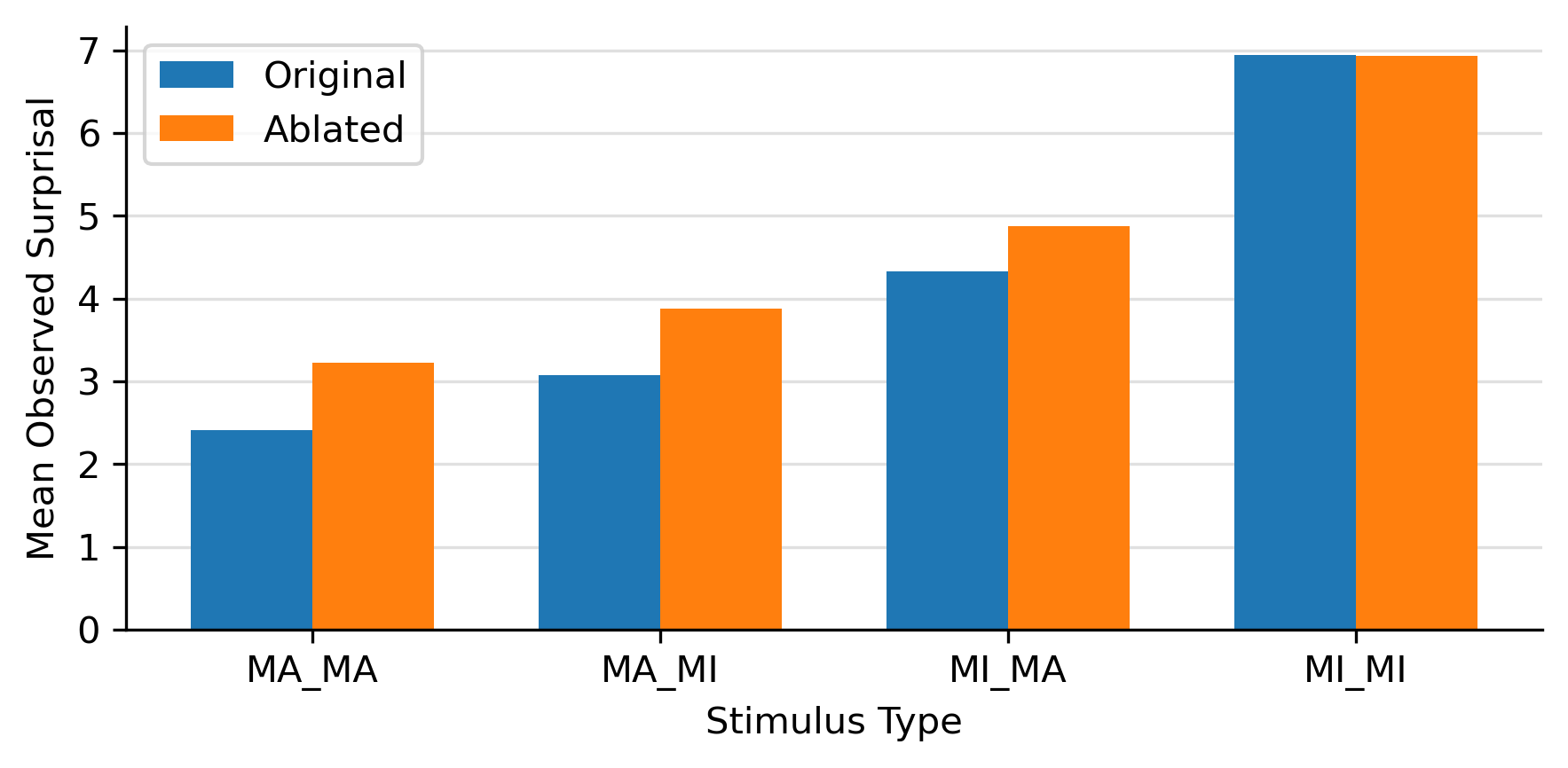}
  \caption{Surprisal in the original vs.\ ablated model.}
  \label{fig:ablation_again1}
\end{figure}

Having analyzed surprisal, attention entropy and energy, we now turn to identifying the contribution of heads in processing discourses, allowing us to find causal components. Using attention entropy from GPT-2 medium, we find heads that diverge across conditions in the `Again' Study 1 from Section~\ref{attent}.  %
We locate the heads (N = 5, 10, 25, 100) with the highest absolute difference between MIMI and MIMA. Then, we turn off these heads by applying value zeroing to identify their contributions to the context, while keeping attention intact \citep{mohebbi-etal-2023-quantifying}, and recalculate surprisal. At 25 heads, MIMI stays the same, but it affects the other conditions as depicted in Figure~\ref{fig:ablation_again1}. This indicates that the heads are likely responsible for tracking presupposition dependencies, including the illusory ones in the coherence illusions. We also \textbf{transfer} the ablation of the same heads to the other experiments. PMIMI and RMIMI from `Again' Study 2 stay relatively the same, while the other conditions get affected, see Figure~\ref{fig:ablation_transfer_again2}. All conditions get affected in `Too' Study 1, see Figure~\ref{fig:ablation_transfer_too1}. These findings indicate that a mechanism, shared across these heads architecturally, underlies the encoding of coherence (both true and illusory), but the encoding is possibly lexically specific and might differ per presupposition trigger.

\section{Probing Hidden Representations}
\begin{figure}[h]
    \centering
    \includegraphics[width=0.9\linewidth]{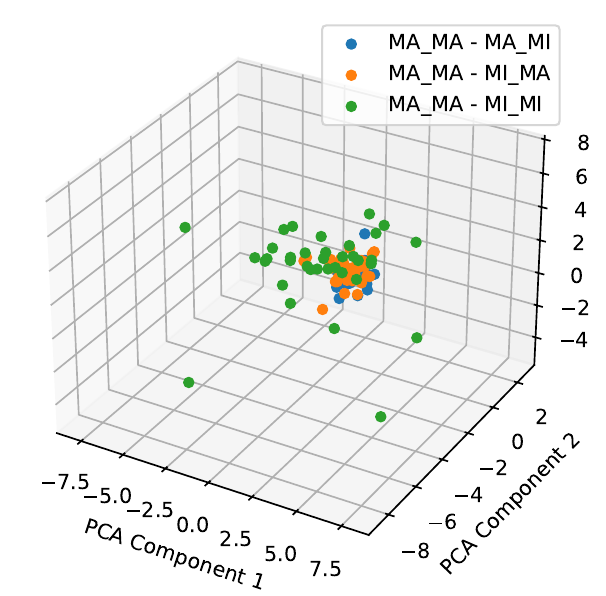}
  \caption{Representations from GPT-2 medium for `Again' Study 1. Difference from MAMA, when the critical word was seen.}
  \label{fig:pcadiff}
\end{figure}

Finally, we analyze whether we can extract information about the coherence conditions from the hidden representations in the medium Dutch GPT-2 model. We extract the final-layer hidden states from the position of the trigger and  critical words. At the trigger position, the representations within stimulus sets (4 conditions) do not markedly differ from each other, see Figure~\ref{fig:tsne}. This is because the discourses \textit{within} a stimulus set are semantically very close to each other; whereas the difference is more visible \textit{between} stimulus sets. We also illustrate the difference between MAMA, the coherent, matching condition and the others in Figure~\ref{fig:pcadiff}, which reveals the dispersion of representations as the context becomes more incoherent when the critical word is seen. The coherence illusion case is  close to MAMA, indicating that, representationally, the model is encoding it closer to a truly coherent representation than an incoherent representation.

To examine whether it is possible to predict  coherence conditions, we train linear probes~\citep{probingbelinkov,ettinger-etal-2016-probing, giulianelli-etal-2018-hood}, consisting of a linear layer projecting from the model's hidden dimensions to 4 classes. We use 80\% of the stimulus sets in training and 20\% to report the findings. We use 10 random seeds to split the dataset and train the probe. We use batch size 4, learning rate 2e-3 (after testing 1e-2, 1e-3, 2e-3, 2e-5), 40 epochs. The best result on the test set is achieved at epoch 31, with 71.88\% accuracy (random baseline: 25\%, 4 classes). The errors are mostly due to MAMA and MAMI being confused with MIMA, supporting our previous findings on the coherence illusion. Although linear probes might have limitations~\citep{probingbelinkov}, our results suggest that models keep track of the discourse quite well, and interestingly, their occasional failures resonate with  coherence illusion.

\section{Conclusion}
Our work connects coherence illusions in human discourse processing with contextual entrainment in LLMs, revealing not only behavioral signals, but also internal states and metrics related to discourse coherence in LLMs in Dutch. We investigated whether LLMs exhibit behavior similar to coherence illusion observed in psycholinguistics studies in humans in Dutch. Utilizing Dutch language models and multilingual LLMs, we found that models prefer matching and coherent discourses, and they are sensitive to distractors in the narrative. Our results show that coherence illusions arise in LLMs exposed to Dutch, and the surprisal results match human behavior. Additionally, we explored attention entropy as a  measure to find  heads in the models that might be relevant for tracking discourse coherence across experimental conditions. We also introduced energy as a novel measure to quantify coherence, inspired by associative memory networks. Probing the representations in the models also revealed that coherence information can be extracted from model internals, showing that the effects of coherence illusions can be analyzed at multiple levels. Regarding language processing accounts, both approaches are revealing: surprisal accounts provide a link to human processing data, which is supported in our study by the observed significant link between surprisal and reading-time data, while memory-based accounts provide insights into what to investigate inside LLMs as a culprit of coherence effects. Our research thus connects both lines of investigations in psycholinguistics, which are sometimes treated as independent or opposing each other, and shows that they can both be useful in approaching one phenomenon. Taken together, our experiments have shown that signs of coherence illusions can be found in model behavior and internals, which can inform future studies on the vulnerability of LLMs to the contents of the prompts as well as on the models of language processing in humans.

\section*{Limitations}

We investigate coherence illusion in Dutch. Our study is limited by the resources for Dutch (models and how good they perform in Dutch). Still, we believe that the set of models we chose represent models that can handle Dutch well. Additionally, we utilize stimuli prepared for psycholinguistic studies on humans, which might have different effects in LLMs. We do not study the effects in multiple languages; however, it would be very interesting to study coherence illusions in other languages and with more samples in the future to see how our results generalize.

\section*{Ethics Statement}
Our goal in this paper was to study the linguistic processes in LLMs concerned with coherence in Dutch discourses. To this end, we used an eye-tracking dataset for which the authors had obtained informed consent from participants and ethical approval before collecting the data. The data does not involve any information revealing participant identities and poses no risk to them. There could be adversarial effects of prompt manipulation and contextual distractors in LLMs, which is beyond the scope of this paper.

\section*{Acknowledgments}
The research reported in this paper was supported by the European Research Council (ERC), grant 101088098 - MEMLANG. Views and opinions expressed are however those of the authors only and do not necessarily reflect those of the European Union or the European Research Council Executive Agency. Neither the European Union nor the granting authority can be held responsible for them. We thank the members of the MEMLANG group for their comments. We used AI assistants for help with some coding tasks and for minor text editing.

\bibliography{custom}

\appendix

\section{Samples from the `Too' Study}
\label{app_too}
An example of an item, translated from Dutch. For further details, see \cite{SCHMITZ2025104637}.\\

\noindent\textbf{MAMA} has a discourse with the target and distractor both matching the critical word. 

\ex. Rutger is \textbf{a businessman}, but Siem is \textbf{not a businessman}. They have
an acquaintance who is \textbf{also a businessman} and earns a lot of money with his company.

\noindent\textbf{MIMA} has a target word mismatching the critical word. 

\ex. Roxanne is \textbf{a businesswoman}, but Siem is \textbf{not a businessman}. They have
an acquaintance \textbf{who is also a businessman} and earns a lot of money with his company.

\noindent\textbf{MAMI} has a distractor mismatching the critical word. 

\ex. Rutger is \textbf{a businessman}, but Susan is \textbf{not a businesswoman}. They have
an acquaintance who is \textbf{also a businessman} and earns a lot of money with his company.

\noindent\textbf{MIMI} has both the target and the distractor mismatching the critical word.

\ex. Roxanne is \textbf{a businesswoman}, but Susan is \textbf{not a businesswoman}. They have
an acquaintance who is \textbf{also a businessman} and earns a lot of money with his company.

\section{Preprocessing and Experimental Setup}
\label{expsetup}

\noindent\textbf{Processing of texts.} The texts contain new line characters as they were shown to the participants as a paragraph, rather than a single line. We remove the new lines and encode the processed text using the tokenizers of the models. When we determine the location of the critical word, depending on the way each model tokenizes, we either add a space at the beginning of words or not.

\noindent\textbf{Experimental setup.} We run the small models locally on a CPU and the LLMs on A100 GPUs. We feed each sentence once to each model as we do not sample from the vocabulary.

\section{Further Surprisal Results}
\begin{figure}[h]
    \centering
    \includegraphics[width=1\linewidth]{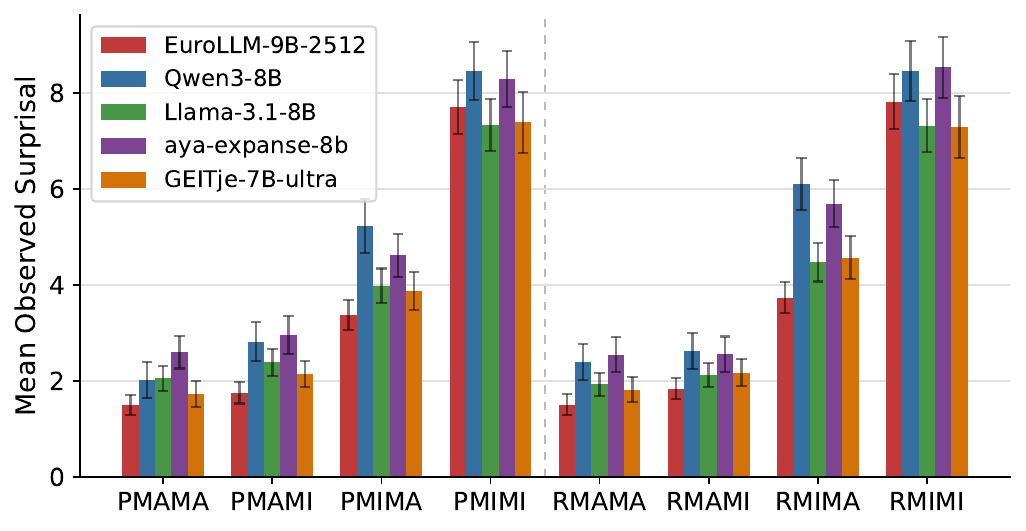}
  \caption{Surprisal results for the `Again' Study 2 for the LLMs.}
  \label{fig:supexp2opnieuwllms}
\end{figure}

\begin{figure}[h]
    \centering
    \includegraphics[width=1\linewidth]{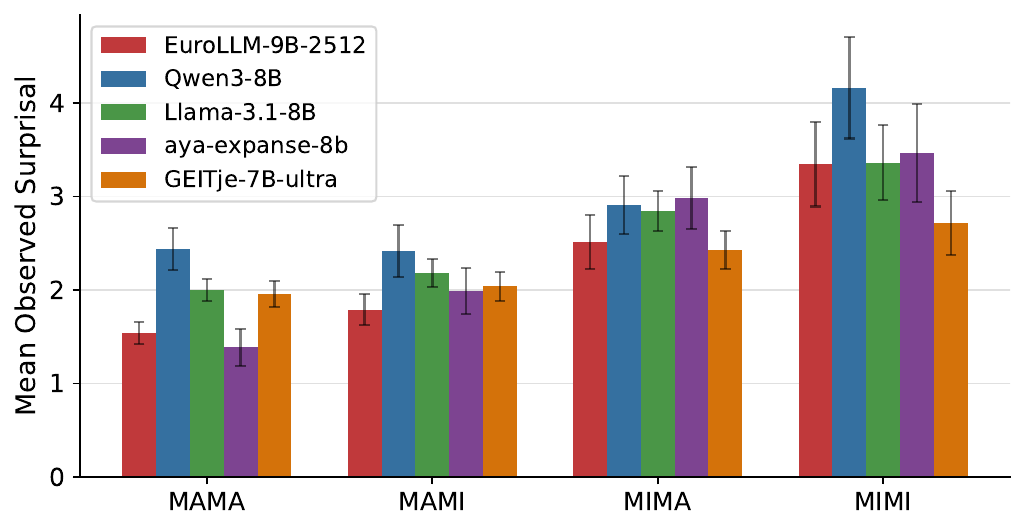}
  \caption{Surprisal results for the `Too' Study  for the LLMs.}
  \label{fig:supexp1toollms}
\end{figure}

We provide additional figures depicting models' surprisals across studies and coherence conditions. Figure~\ref{fig:supexp2opnieuwllms} visualizes the surprisal  results for the `Again' Study 2 for the LLMs, and Figure~\ref{fig:supexp1toollms} provides the surprisal results for the `Too' Study  for the LLMs.

\section{Attention Entropy Matrices}
\label{app_ent}
\begin{figure}[h]
    \centering
    \includegraphics[width=0.78\linewidth]{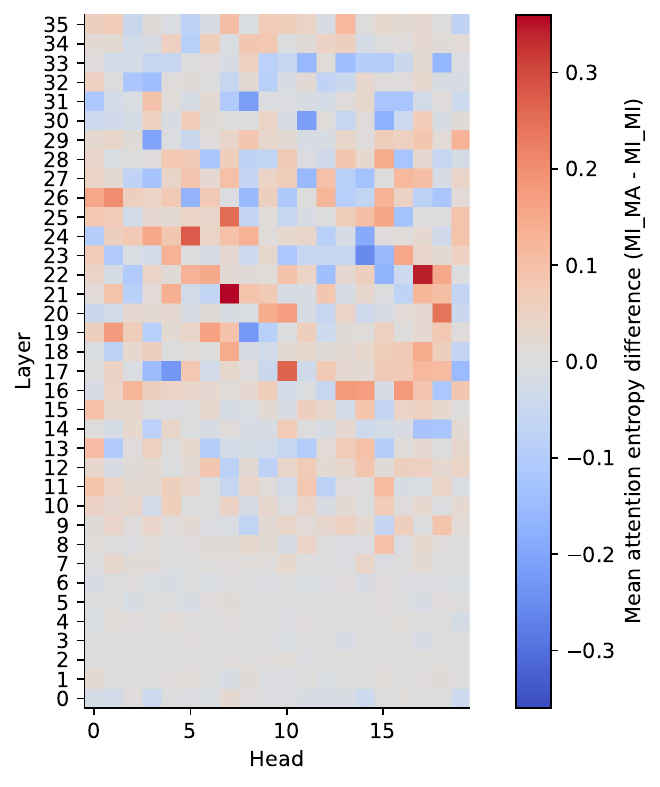}
  \caption{Attention Entropy difference in `Again' Study 1 between MIMA and MIMI in the large Dutch GPT-2 model's heads across layers.}
  \label{fig:attentrdifflarge}
\end{figure}

\begin{figure}[h]
    \centering
    \includegraphics[width=0.8\linewidth]{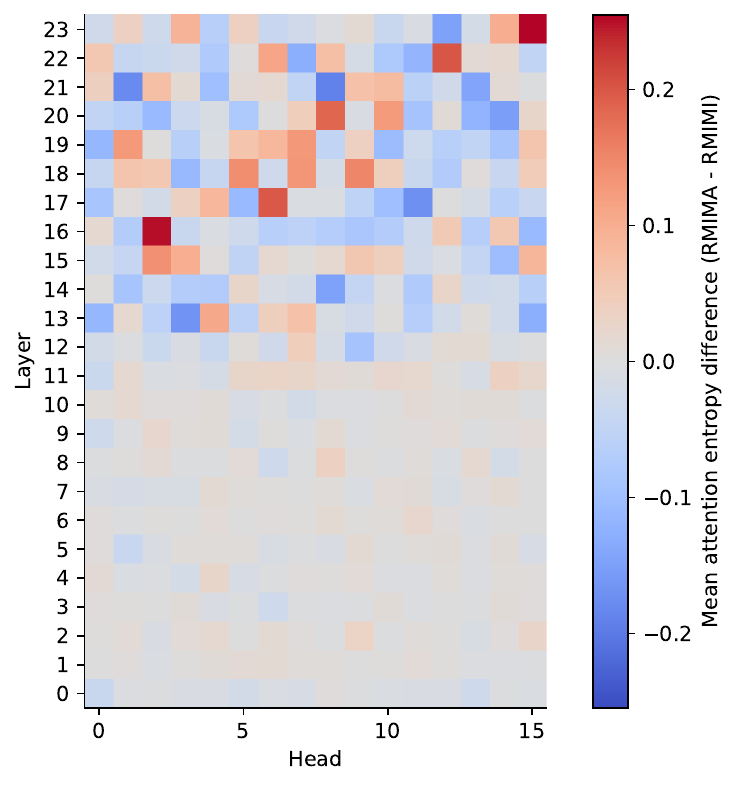}
  \caption{Attention Entropy difference in `Again' Study 2 between RMIMA and RMIMI in the medium Dutch GPT-2 model's heads across layers.}
  \label{fig:attentrdiffmediumPR}
\end{figure}

\begin{figure}[h]
    \centering
    \includegraphics[width=0.8\linewidth]{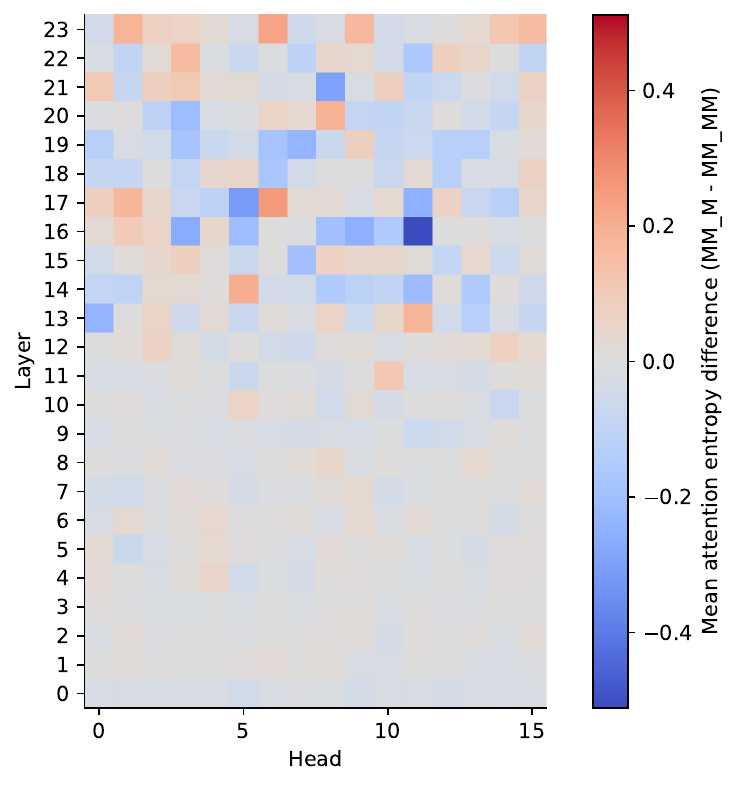}
  \caption{Attention Entropy difference in `Too' Study between MIMA and MIMI in the medium Dutch GPT-2 model's heads across layers.}
  \label{fig:attentrdiffmediumTOOMM}
\end{figure}

\begin{figure}[h]
    \centering
    \includegraphics[width=0.8\linewidth]{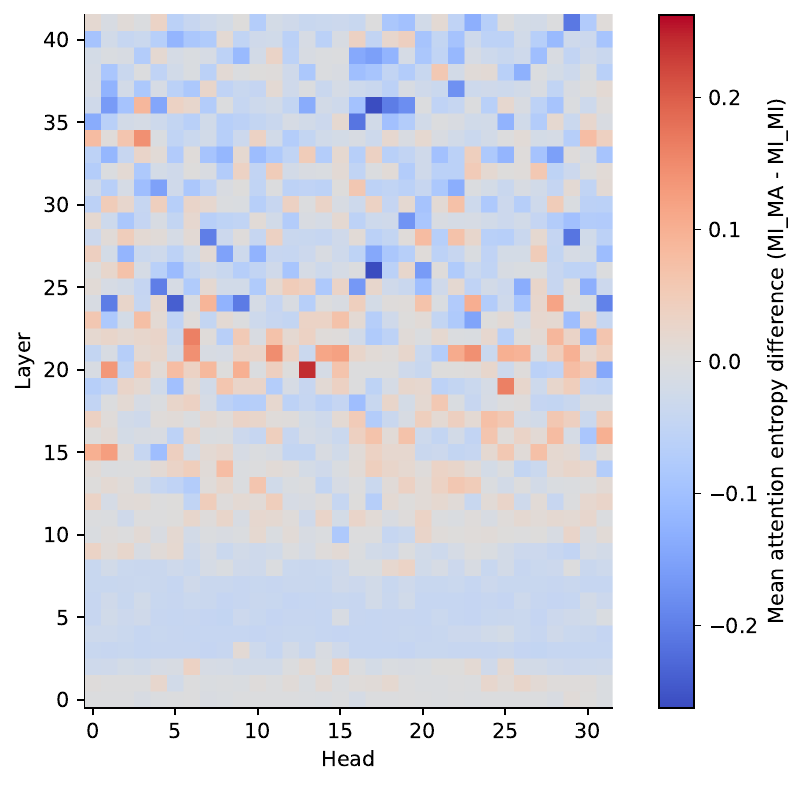}
  \caption{Attention Entropy difference in `Again' Study 1 between MIMA and MIMI in EuroLLM's heads across layers.}
  \label{fig:attentrdiffeuro}
\end{figure}

We provide additional matrices depicting attention entropy differences in more studies and model types in Figures~\ref{fig:attentrdifflarge},~\ref{fig:attentrdiffmediumPR}, ~\ref{fig:attentrdiffmediumTOOMM} and ~\ref{fig:attentrdiffeuro}.

\section{Energy Scores for GPT-2-Large}
\begin{figure}[h]
    \centering
    \includegraphics[width=0.96\linewidth]{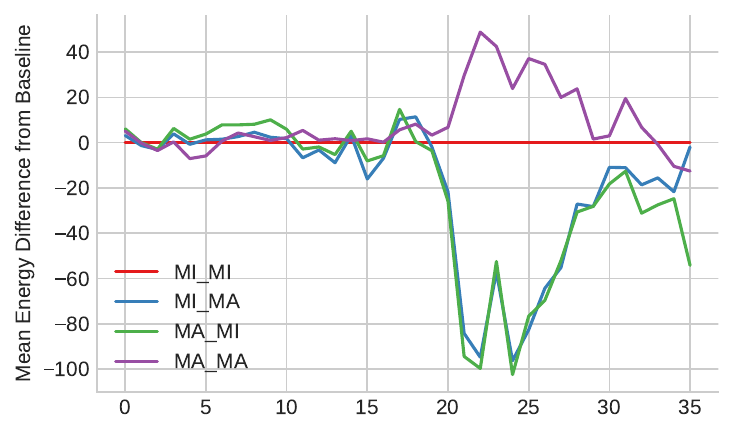}
  \caption{\textbf{Energy 1} scores across layers for the \textit{large} Dutch GPT-2  model for the `Again' Study 1.}
  \label{fig:energy1exp1large}
\end{figure}

\begin{figure}[h]
    \centering
    \includegraphics[width=0.96\linewidth]{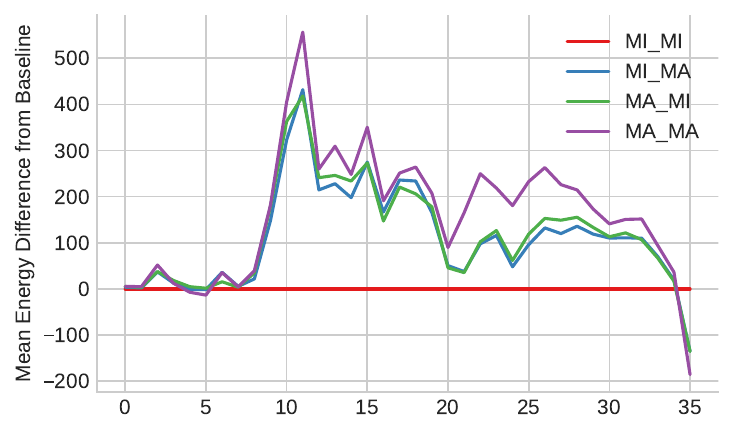}
  \caption{\textbf{Energy 2} scores across layers for the \textit{large} Dutch GPT-2  model for the `Again' Study 1.}
  \label{fig:energy2exp1large}
\end{figure}

Figure~\ref{fig:energy1exp1large} and \ref{fig:energy2exp1large} depict the energy scores for the GPT-2 large model, Energy 1 and Energy 2, respectively. The patterns are generally aligned with those of the medium model.

\section{Head Ablation Transfer}

\begin{figure}[h]
    \centering
    \includegraphics[width=1\linewidth]{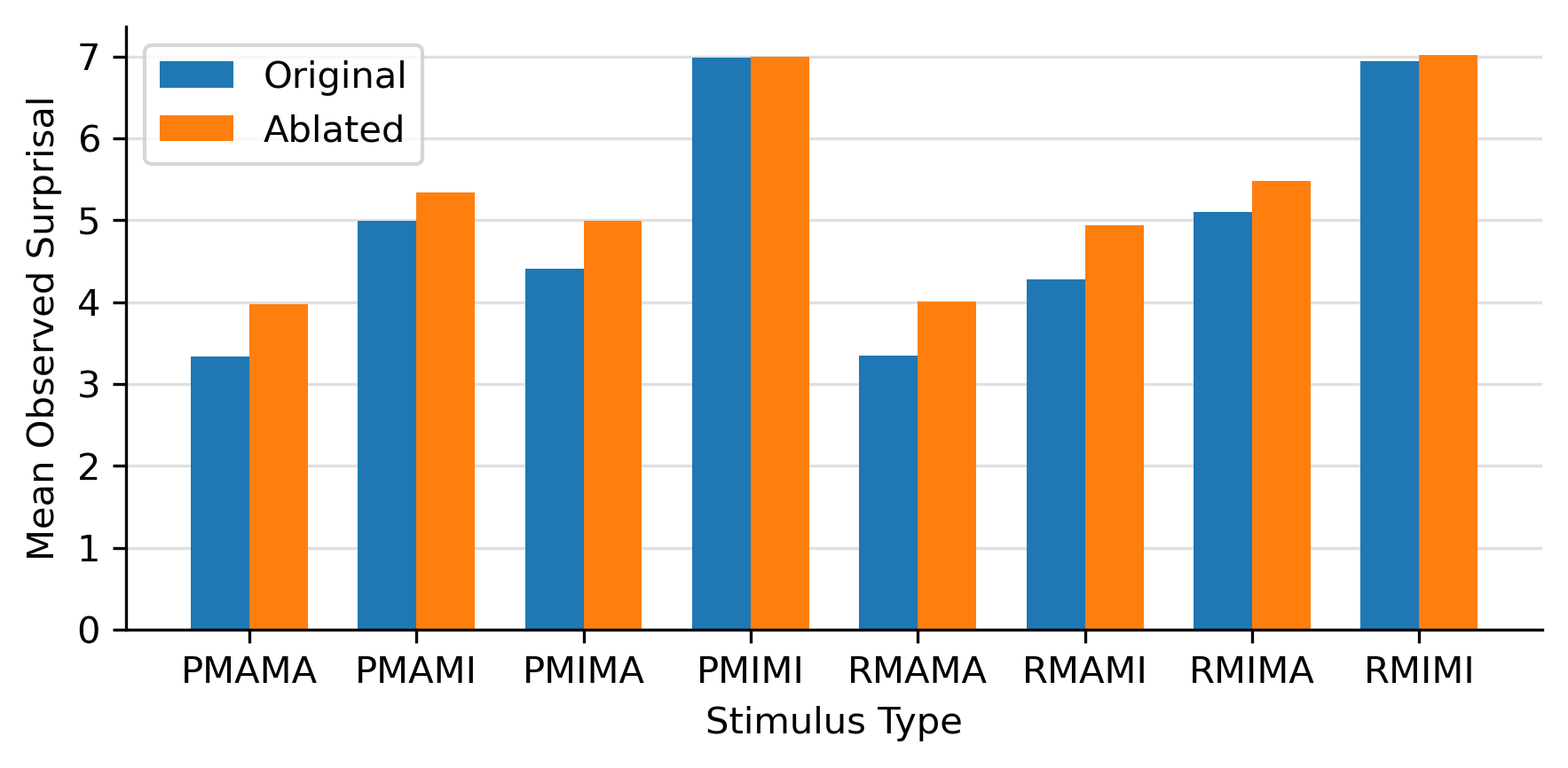}
  \caption{Surprisal after transferring head ablation to `Again' Study 2.}
  \label{fig:ablation_transfer_again2}
\end{figure}

\begin{figure}[h!]
    \centering
    \includegraphics[width=1\linewidth]{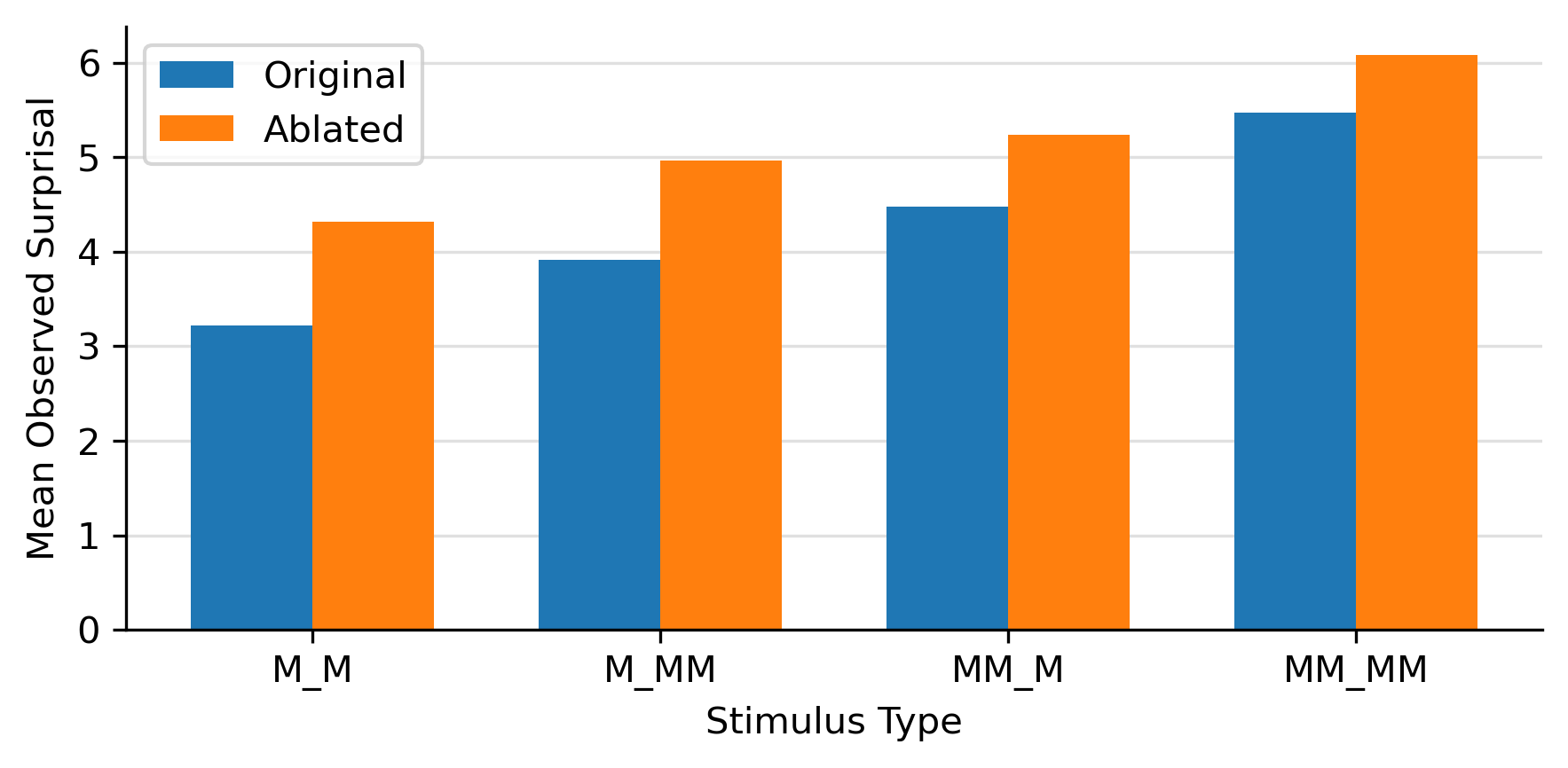}
  \caption{Surprisal after transferring head ablation to `Too' Study.}
  \label{fig:ablation_transfer_too1}
\end{figure}

Figures~\ref{fig:ablation_transfer_again2} and ~\ref{fig:ablation_transfer_too1} illustrate the changes in surprisal after the values of 25 heads were turned off in transfer studies.

\section{Representational Space}
\begin{figure}[h!]
    \centering
    \includegraphics[width=0.9\linewidth]{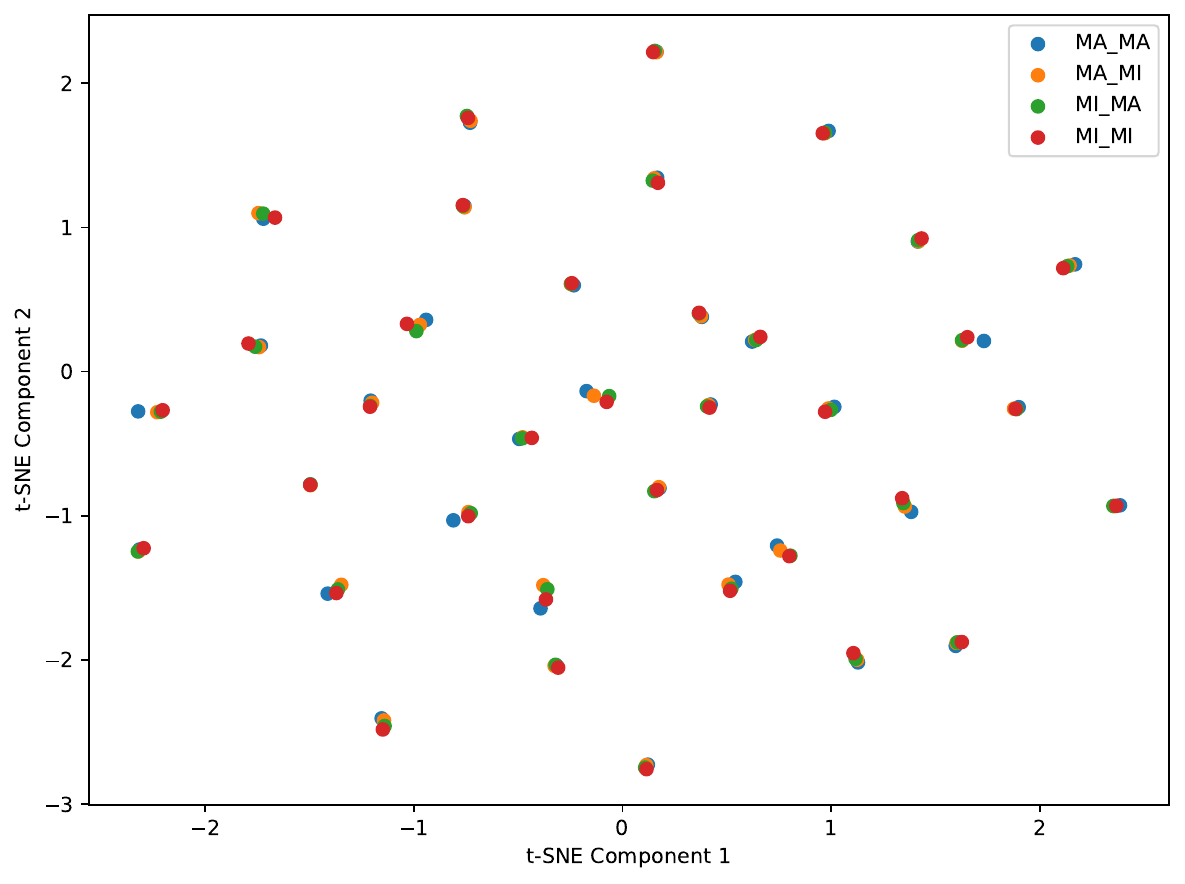}
  \caption{Representations from GPT-2 medium for `Again' Study 1 at the trigger position.}
  \label{fig:tsne}
\end{figure}

Figure~\ref{fig:tsne} depicts the representations from GPT-2 medium for `Again' Study 1 at the trigger position.

\section{Packages Used}
We use Python version 3.13, PyTorch version 2.7, HuggingFace version 4.57.

\section{Further Results}
\label{app:lmem-more}

Surprisal-only models, see Table~\ref{tab:lmem_results_simple}, `Again' Study 1:

\textbf{Pre-critical region}: Surprisal is significant only for the TOTFIX in the GEITje model measure, possibly due to LLM regressing to resolve the presupposition. 

\textbf{Critical region}: Surprisal is significant across measures for both models.

\textbf{Post-Critical / Spillover region}: Only TGDUR measure's Surprisal is significant for both models, suggesting some spillover of the processing difficulty, but that does not generalize across measures.

`Again' Study 2:

\textbf{Pre-critical region}: Surprisal is significant across measures for both models.

\textbf{Critical region}: Surprisal is significant across measures for both models.

\textbf{Post-Critical / Spillover region}: TGDUR and TOTFIXDUR measure's Surprisal is significant, suggesting some spillover of the processing difficulty, but that does not generalize across all measures and does not hold for both models.

Surprisal, Target and Distractor models, see Table~\ref{tab:lmem_results}, `Again' Study 1:

\textbf{Pre-critical region}: Surprisal is significant only for the TOTFIX measure, possibly due to LLM regressing to resolve the presupposition. 

\textbf{Critical region}: Surprisal is significant, but only in the GPT-Neo model. The lack of significance compared to the surprisal-only study suggests that the surprisal effect captures and is closely related to the target and distractor match and mismatch manipulation.

\textbf{Post-Critical / Spillover region}: Only TG measure's Surprisal is significant.

`Again' Study 2:

\textbf{Pre-critical region}: Surprisal is significant across measures.

\textbf{Critical region}: Surprisal is significant, but only in TGDUR for both models. The lack of significance for FFDUR and TOTFIXDUR compared to the surprisal-only study suggests that also here, the surprisal effect captures and is closely related to the target and distractor match and mismatch manipulation.

\textbf{Post-Critical / Spillover region}: Only TG measure's Surprisal is significant in the GEITje model.

\begin{table*}[htbp]
\centering
\resizebox{\linewidth}{!}{%
\renewcommand{\arraystretch}{1.1}
\begin{tabular}{l l l l rr rr rr}
\toprule
& & & &
  \multicolumn{2}{c}{\textbf{Pre-Critical}} &
  \multicolumn{2}{c}{\textbf{Critical}} &
  \multicolumn{2}{c}{\textbf{Post-Critical}} \\
\cmidrule(lr){5-6}\cmidrule(lr){7-8}\cmidrule(lr){9-10}
\textbf{Model} & \textbf{Exp.} & \textbf{Metric} & \textbf{Measure} &
  \textbf{Int.} & \textbf{Surp.} &
  \textbf{Int.} & \textbf{Surp.} &
  \textbf{Int.} & \textbf{Surp.} \\
\midrule

\multirow{12}{*}{\shortstack[l]{BramVanroy/\\GEITje-7B-ultra}}
  & \multirow{6}{*}{E1}
    & Value     & ffdur      & 5.265 &  0.018 & 5.347 &  0.015 & 5.332 &  0.010 \\
  & & $p$-value & ffdur      & 0.000 &  0.077 & 0.000 &  0.025 & 0.000 &  0.308 \\
  & & Value     & tgdur      & 5.337 &  0.020 & 6.099 &  0.039 & 5.694 &  0.043 \\
  & & $p$-value & tgdur      & 0.000 &  0.101 & 0.000 &  0.000 & 0.000 &  0.029 \\
  & & Value     & totfixdur  & 5.600 &  0.050 & 6.338 &  0.065 & 5.983 &  0.025 \\
  & & $p$-value & totfixdur  & 0.000 &  0.008 & 0.000 &  0.000 & 0.000 &  0.360 \\
\cmidrule(l){2-10}
  & \multirow{6}{*}{E2}
    & Value     & ffdur      & 5.345 &  0.023 & 5.393 &  0.017 & 5.387 & $-$0.005 \\
  & & $p$-value & ffdur      & 0.000 &  0.043 & 0.000 &  0.010 & 0.000 &  0.580 \\
  & & Value     & tgdur      & 5.457 &  0.062 & 6.250 &  0.084 & 5.864 &  0.053 \\
  & & $p$-value & tgdur      & 0.000 &  0.001 & 0.000 &  0.000 & 0.000 &  0.006 \\
  & & Value     & totfixdur  & 5.849 &  0.062 & 6.641 &  0.114 & 6.125 &  0.047 \\
  & & $p$-value & totfixdur  & 0.000 &  0.027 & 0.000 &  0.000 & 0.000 &  0.046 \\
\midrule

\multirow{12}{*}{\shortstack[l]{yhavinga/\\gpt-neo-\\125M-dutch}}
  & \multirow{6}{*}{E1}
    & Value     & ffdur      & 5.264 & $-$0.008 & 5.344 &  0.018 & 5.332 &  0.018 \\
  & & $p$-value & ffdur      & 0.000 &  0.399   & 0.000 &  0.008 & 0.000 &  0.059 \\
  & & Value     & tgdur      & 5.335 & $-$0.004 & 6.082 &  0.063 & 5.691 &  0.077 \\
  & & $p$-value & tgdur      & 0.000 &  0.745   & 0.000 &  0.000 & 0.000 &  0.000 \\
  & & Value     & totfixdur  & 5.596 &  0.009   & 6.319 &  0.099 & 5.976 &  0.092 \\
  & & $p$-value & totfixdur  & 0.000 &  0.647   & 0.000 &  0.000 & 0.000 &  0.001 \\
\cmidrule(l){2-10}
  & \multirow{6}{*}{E2}
    & Value     & ffdur      & 5.346 &  0.041   & 5.393 &  0.011 & 5.387 & $-$0.011 \\
  & & $p$-value & ffdur      & 0.000 &  0.000   & 0.000 &  0.103 & 0.000 &  0.187 \\
  & & Value     & tgdur      & 5.458 &  0.069   & 6.250 &  0.107 & 5.864 &  0.033 \\
  & & $p$-value & tgdur      & 0.000 &  0.000   & 0.000 &  0.000 & 0.000 &  0.102 \\
  & & Value     & totfixdur  & 5.851 &  0.091   & 6.641 &  0.130 & 6.125 &  0.029 \\
  & & $p$-value & totfixdur  & 0.000 &  0.001   & 0.000 &  0.000 & 0.000 &  0.252 \\

\bottomrule
\end{tabular}
}%
\caption{%
  LMEM results per region.
  Model structure:
  \texttt{log(\_dur) $\sim$ scale(surprisal) + (1\,|\,subj) + (1\,|\,item)}.
  Columns show fixed-effect estimates and $p$-values
  for the intercept and surprisal predictor for three regions.
}
\label{tab:lmem_results_simple}
\end{table*}

\begin{table*}[htbp]
\centering
\resizebox{\linewidth}{!}{%
\renewcommand{\arraystretch}{1.1}
\begin{tabular}{%
  l l l l
  r r r r r
  r r r r r
  r r r r r
}
\toprule
& & & &
  \multicolumn{5}{c}{\textbf{Pre-Critical Region}} &
  \multicolumn{5}{c}{\textbf{Critical Region}} &
  \multicolumn{5}{c}{\textbf{Post-Critical Region}} \\
\cmidrule(lr){5-9}\cmidrule(lr){10-14}\cmidrule(lr){15-19}
\textbf{Model} & \textbf{Exp.} & \textbf{Metric} & \textbf{Measure} &
  \textbf{Int.} & \textbf{Surp.} & \textbf{Tar.} & \textbf{Dis.} & \textbf{Tar:Dis} &
  \textbf{Int.} & \textbf{Surp.} & \textbf{Tar.} & \textbf{Dis.} & \textbf{Tar:Dis} &
  \textbf{Int.} & \textbf{Surp.} & \textbf{Tar.} & \textbf{Dis.} & \textbf{Tar:Dis} \\
\midrule

\multirow{12}{*}{\shortstack[l]{BramVanroy/\\GEITje-7B-ultra}}
  & \multirow{6}{*}{E1}
    & Value      & ffdur
      & 5.265 & 0.018  &  0.007  & $-$0.004 &  0.005
      & 5.347 & 0.000  & $-$0.018 & $-$0.002 & $-$0.001
      & 5.332 & 0.013  & $-$0.014 &  0.009  &  0.008  \\
  & & $p$-value  & ffdur
      & 0.000 & 0.070  &  0.393  &  0.638  &  0.543
      & 0.000 & 0.977  &  0.167  &  0.791  &  0.909
      & 0.000 & 0.210  &  0.100  &  0.250  &  0.308  \\
  & & Value      & tgdur
      & 5.337 & 0.019  & $-$0.003 & $-$0.001 & $-$0.015
      & 6.099 & 0.031  & $-$0.009 & $-$0.010 & $-$0.010
      & 5.694 & 0.068  & $-$0.071 &  0.001  & $-$0.011 \\
  & & $p$-value  & tgdur
      & 0.000 & 0.133  &  0.793  &  0.886  &  0.129
      & 0.000 & 0.183  &  0.630  &  0.408  &  0.353
      & 0.000 & 0.001  &  0.000  &  0.946  &  0.310  \\
  & & Value      & totfixdur
      & 5.599 & 0.050  & $-$0.065 & $-$0.002 & $-$0.008
      & 6.338 & 0.012  & $-$0.061 & $-$0.013 & $-$0.006
      & 5.983 & 0.064  & $-$0.072 & $-$0.011 & $-$0.010 \\
  & & $p$-value  & totfixdur
      & 0.000 & 0.009  &  0.000  &  0.910  &  0.561
      & 0.000 & 0.689  &  0.016  &  0.429  &  0.658
      & 0.000 & 0.023  &  0.000  &  0.406  &  0.438  \\
\cmidrule(l){2-19}
  & \multirow{6}{*}{E2}
    & Value      & ffdur
      & 5.345 &  0.023 &  0.015  & $-$0.007 & $-$0.003
      & 5.393 &  0.005 & $-$0.015 & $-$0.007 &  0.000
      & 5.387 & $-$0.007 & $-$0.025 & $-$0.020 &  0.018 \\
  & & $p$-value  & ffdur
      & 0.000 &  0.042 &  0.051  &  0.350  &  0.708
      & 0.000 &  0.596 &  0.102  &  0.308  &  0.977
      & 0.000 &  0.406 &  0.002  &  0.015  &  0.027  \\
  & & Value      & tgdur
      & 5.457 &  0.063 &  0.024  & $-$0.009 &  0.001
      & 6.250 &  0.064 & $-$0.026 & $-$0.015 & $-$0.026
      & 5.864 &  0.048 & $-$0.050 & $-$0.018 & $-$0.001 \\
  & & $p$-value  & tgdur
      & 0.000 &  0.001 &  0.016  &  0.362  &  0.933
      & 0.000 &  0.001 &  0.071  &  0.157  &  0.008
      & 0.000 &  0.012 &  0.000  &  0.062  &  0.957  \\
  & & Value      & totfixdur
      & 5.849 &  0.057 & $-$0.043 & $-$0.020 & $-$0.013
      & 6.642 &  0.045 & $-$0.088 & $-$0.021 & $-$0.030
      & 6.126 &  0.044 & $-$0.063 & $-$0.005 & $-$0.007 \\
  & & $p$-value  & totfixdur
      & 0.000 &  0.043 &  0.001  &  0.138  &  0.326
      & 0.000 &  0.062 &  0.000  &  0.124  &  0.017
      & 0.000 &  0.064 &  0.000  &  0.643  &  0.527  \\
\midrule

\multirow{12}{*}{\shortstack[l]{yhavinga/\\gpt-neo-\\125M-dutch}}
  & \multirow{6}{*}{E1}
    & Value      & ffdur
      & 5.264 & $-$0.008 &  0.009  & $-$0.004 &  0.001
      & 5.344 &  0.011  & $-$0.014 & $-$0.001 & $-$0.001
      & 5.332 &  0.018  & $-$0.012 &  0.006  &  0.008  \\
  & & $p$-value  & ffdur
      & 0.000 &  0.400  &  0.271  &  0.645  &  0.897
      & 0.000 &  0.242  &  0.077  &  0.896  &  0.858
      & 0.000 &  0.061  &  0.146  &  0.418  &  0.305  \\
  & & Value      & tgdur
      & 5.336 & $-$0.004 & $-$0.002 &  0.000  & $-$0.019
      & 6.082 &  0.064  & $-$0.005 & $-$0.002 & $-$0.015
      & 5.691 &  0.075  & $-$0.060 & $-$0.005 & $-$0.008 \\
  & & $p$-value  & tgdur
      & 0.000 &  0.734  &  0.844  &  0.976  &  0.050
      & 0.000 &  0.002  &  0.708  &  0.850  &  0.107
      & 0.000 &  0.000  &  0.000  &  0.627  &  0.450  \\
  & & Value      & totfixdur
      & 5.595 &  0.007  & $-$0.064 & $-$0.002 & $-$0.018
      & 6.319 &  0.086  & $-$0.034 &  0.008  & $-$0.024
      & 5.976 &  0.087  & $-$0.064 & $-$0.013 & $-$0.006 \\
  & & $p$-value  & totfixdur
      & 0.000 &  0.699  &  0.000  &  0.884  &  0.195
      & 0.000 &  0.001  &  0.036  &  0.553  &  0.044
      & 0.000 &  0.002  &  0.000  &  0.292  &  0.652  \\
\cmidrule(l){2-19}
  & \multirow{6}{*}{E2}
    & Value      & ffdur
      & 5.346 &  0.040  &  0.015  & $-$0.007 & $-$0.004
      & 5.393 &  0.000  & $-$0.018 & $-$0.009 &  0.002
      & 5.387 & $-$0.012 & $-$0.025 & $-$0.019 &  0.018  \\
  & & $p$-value  & ffdur
      & 0.000 &  0.000  &  0.061  &  0.348  &  0.593
      & 0.000 &  0.990  &  0.013  &  0.196  &  0.813
      & 0.000 &  0.149  &  0.002  &  0.016  &  0.028  \\
  & & Value      & tgdur
      & 5.458 &  0.068  &  0.023  & $-$0.009 & $-$0.001
      & 6.250 &  0.094  & $-$0.030 & $-$0.002 & $-$0.027
      & 5.864 &  0.029  & $-$0.050 & $-$0.020 &  0.000  \\
  & & $p$-value  & tgdur
      & 0.000 &  0.000  &  0.021  &  0.346  &  0.909
      & 0.000 &  0.000  &  0.006  &  0.853  &  0.003
      & 0.000 &  0.148  &  0.000  &  0.036  &  0.988  \\
  & & Value      & totfixdur
      & 5.851 &  0.090  & $-$0.043 & $-$0.020 & $-$0.015
      & 6.642 &  0.081  & $-$0.084 & $-$0.006 & $-$0.033
      & 6.126 &  0.025  & $-$0.063 & $-$0.007 & $-$0.007 \\
  & & $p$-value  & totfixdur
      & 0.000 &  0.001  &  0.001  &  0.137  &  0.276
      & 0.000 &  0.000  &  0.000  &  0.661  &  0.004
      & 0.000 &  0.327  &  0.000  &  0.522  &  0.563  \\

\bottomrule
\end{tabular}
}%
\caption{%
  LMEM results per region.
  Model structure:
  \texttt{log(\_dur) $\sim$ scale(surprisal) + target + distractor
  + target\,$\times$\,distractor + (1\,|\,subj) + (1\,|\,item)}.
  Columns show fixed-effect estimates and $p$-values per predictor
  for three regions.
  \textit{Int.}$=$Intercept; \textit{Surp.}$=$Surprisal;
  \textit{Tar.}$=$Target; \textit{Dis.}$=$Distractor;
  \textit{Tar:Dis}$=$interaction.
}
\label{tab:lmem_results}
\end{table*}

\begin{table*}[htbp]
\centering
\resizebox{\linewidth}{!}{%
\renewcommand{\arraystretch}{1.1}
\begin{tabular}{l l l l c c c c}
\toprule
\textbf{Model} & \textbf{Exp.} & \textbf{Surprisal} & \textbf{Type} &
  \textbf{Pre-Critical} & \textbf{Critical} & \textbf{Post-Critical} & \textbf{Object} \\
\midrule
\multirow{16}{*}{\shortstack[l]{BramVanroy/\\GEITje-7B-ultra}}
  & \multirow{8}{*}{1}
    & \multirow{4}{*}{Total}
      & MA\_MA & 9.400 (0.391)  & 5.059 (0.311)  & 21.741 (0.930) & 3.554 (0.363)  \\
  & & & MA\_MI & 9.778 (0.370)  & 6.242 (0.400)  & 21.896 (0.797) & 4.677 (0.472)  \\
  & & & MI\_MA & 9.761 (0.367)  & 10.686 (0.443) & 20.714 (0.814) & 8.538 (0.407)  \\
  & & & MI\_MI & 9.365 (0.372)  & 19.449 (0.782) & 21.003 (0.809) & 17.393 (0.727) \\
\cmidrule(l){3-8}
  & & \multirow{4}{*}{Mean}
      & MA\_MA & 9.400 (0.391)  & 1.639 (0.096)  & 7.247 (0.310)  & 1.690 (0.162) \\
  & & & MA\_MI & 9.778 (0.370)  & 2.032 (0.132)  & 7.299 (0.266)  & 2.249 (0.228) \\
  & & & MI\_MA & 9.761 (0.367)  & 3.496 (0.154)  & 6.905 (0.271)  & 4.159 (0.209) \\
  & & & MI\_MI & 9.365 (0.372)  & 6.350 (0.273)  & 7.001 (0.270)  & 8.441 (0.381) \\
\cmidrule(l){2-8}
  & \multirow{8}{*}{2}
    & \multirow{4}{*}{Total}
      & MA\_MA & 12.389 (0.338) & 5.751 (0.371)  & 20.046 (0.616) & 4.120 (0.306)  \\
  & & & MA\_MI & 12.562 (0.327) & 6.684 (0.411)  & 20.401 (0.596) & 4.945 (0.330)  \\
  & & & MI\_MA & 12.636 (0.334) & 10.383 (0.474) & 19.968 (0.570) & 8.516 (0.375)  \\
  & & & MI\_MI & 12.534 (0.337) & 18.083 (0.765) & 20.619 (0.553) & 15.732 (0.668) \\
\cmidrule(l){3-8}
  & & \multirow{4}{*}{Mean}
      & MA\_MA & 12.389 (0.338) & 1.882 (0.124)  & 6.682 (0.205)  & 1.999 (0.151) \\
  & & & MA\_MI & 12.562 (0.327) & 2.185 (0.136)  & 6.800 (0.199)  & 2.397 (0.161) \\
  & & & MI\_MA & 12.636 (0.334) & 3.389 (0.159)  & 6.656 (0.190)  & 4.125 (0.187) \\
  & & & MI\_MI & 12.534 (0.337) & 5.897 (0.257)  & 6.873 (0.184)  & 7.617 (0.335) \\
\midrule
\multirow{16}{*}{\shortstack[l]{yhavinga/\\gpt-neo-\\125M-dutch}}
  & \multirow{8}{*}{1}
    & \multirow{4}{*}{Total}
      & MA\_MA & 9.164 (0.239) & 7.792 (0.508)  & 19.527 (0.682) & 5.583 (0.463)  \\
  & & & MA\_MI & 9.188 (0.233) & 9.436 (0.629)  & 19.550 (0.679) & 7.145 (0.552)  \\
  & & & MI\_MA & 9.198 (0.236) & 10.416 (0.594) & 19.593 (0.693) & 8.053 (0.519)  \\
  & & & MI\_MI & 9.206 (0.231) & 16.406 (0.985) & 19.707 (0.691) & 13.282 (0.826) \\
\cmidrule(l){3-8}
  & & \multirow{4}{*}{Mean}
      & MA\_MA & 9.164 (0.239) & 2.530 (0.163)  & 6.509 (0.227)  & 2.684 (0.223) \\
  & & & MA\_MI & 9.188 (0.233) & 3.066 (0.204)  & 6.517 (0.226)  & 3.446 (0.269) \\
  & & & MI\_MA & 9.198 (0.236) & 3.388 (0.193)  & 6.531 (0.231)  & 3.894 (0.257) \\
  & & & MI\_MI & 9.206 (0.231) & 5.353 (0.330)  & 6.569 (0.230)  & 6.455 (0.421) \\
\cmidrule(l){2-8}
  & \multirow{8}{*}{2}
    & \multirow{4}{*}{Total}
      & MA\_MA & 9.968 (0.231)  & 9.757 (0.362)  & 19.905 (0.398) & 7.894 (0.372)  \\
  & & & MA\_MI & 10.015 (0.228) & 11.399 (0.373) & 19.955 (0.400) & 9.405 (0.379)  \\
  & & & MI\_MA & 10.022 (0.229) & 11.534 (0.378) & 19.955 (0.399) & 9.541 (0.383)  \\
  & & & MI\_MI & 10.036 (0.228) & 16.758 (0.690) & 20.048 (0.414) & 13.918 (0.622) \\
\cmidrule(l){3-8}
  & & \multirow{4}{*}{Mean}
      & MA\_MA & 9.968 (0.231)  & 3.189 (0.123)  & 6.635 (0.133)  & 3.839 (0.187) \\
  & & & MA\_MI & 10.015 (0.228) & 3.725 (0.127)  & 6.652 (0.133)  & 4.573 (0.193) \\
  & & & MI\_MA & 10.022 (0.229) & 3.767 (0.129)  & 6.652 (0.133)  & 4.636 (0.194) \\
  & & & MI\_MI & 10.036 (0.228) & 5.481 (0.236)  & 6.683 (0.138)  & 6.775 (0.320) \\
\bottomrule
\end{tabular}}
\caption{%
  Surprisals and Standard Errors per region.
}
\label{tab:surprisal_by_region}
\end{table*}

\clearpage
\begin{table}[htb]
\centering 
\begin{tabular}{p{1.6cm}|p{3.8cm}|p{1.2cm}}
\multicolumn{1}{c|}{Region} & 
\multicolumn{1}{c|}{Definition} & 
\multicolumn{1}{c}{Example} \\
\hline

{\footnotesize Pre-critical region} & 
{\footnotesize Only the word \textit{opnieuw} (\textit{again} in English)} & 
{\footnotesize opnieuw} \\ 
\hline

{\footnotesize Critical region} & 
{\footnotesize The region immediately after the word \textit{opnieuw} until the first verb after the word \textit{opnieuw}} & 
{\footnotesize een appel gegeten} \\
\hline

{\footnotesize Post-critical region} & 
{\footnotesize The first three words after the critical region} & 
{\footnotesize want ze wilde} \\ 
\hline

{\footnotesize Object region} & 
{\footnotesize The direct object, including the article in the critical region} & 
{\footnotesize een appel} \\
\hline

\end{tabular}
\caption{Four regions for the eye-tracking analysis.}
\label{fig:four_regions_for_analysis}
\end{table}

\clearpage
    \begin{table*}[htb]
    \centering 
    \begin{tabular}{l | p{11.5cm}} 
    
    \multicolumn{1}{c|}{Type} & \multicolumn{1}{c}{Text} \\
    \hline
    MI\_MI & {\footnotesize Megan had honger maar gelukkig stond er een volle fruitschaal in de lobby van het hotel. Megan heeft een gele peer gegeten, maar ze heeft geen groene peer gegeten. De ochtend erna heeft Megan opnieuw een appel gegeten want ze wilde meer fruit eten.} \\ 
    \hline
    MI\_MA & {\footnotesize Megan had honger maar gelukkig stond er een volle fruitschaal in de lobby van het hotel. Megan heeft een gele peer gegeten, maar ze heeft geen groene appel gegeten. De ochtend erna heeft Megan opnieuw een appel gegeten want ze wilde meer fruit eten.}\\
    \hline
    MA\_MI & {\footnotesize Megan had honger maar gelukkig stond er een volle fruitschaal in de lobby van het hotel. Megan heeft een gele appel gegeten, maar ze heeft geen groene peer gegeten. De ochtend erna heeft Megan opnieuw een appel gegeten want ze wilde meer fruit eten.} \\ 
    \hline
    MA\_MA & {\footnotesize Megan had honger maar gelukkig stond er een volle fruitschaal in de lobby van het hotel. Megan heeft een gele appel gegeten, maar ze heeft geen groene appel gegeten. De ochtend erna heeft Megan opnieuw een appel gegeten want ze wilde meer fruit eten.} \\
    \hline
    \end{tabular}
    \caption{Sample sentences from `Again' Study 1.}
    \label{fig:dutch_discourse_negation}
    \end{table*}

\clearpage
	\begin{table*}[htb]
    \centering 
    \begin{tabular}{l | p{11.5cm}} 
    
    \multicolumn{1}{c|}{Type} & \multicolumn{1}{c}{Text} \\
    \hline
    R\_MI\_MI & {\footnotesize De manager en de assistent haalden in de pauze een stuk fruit uit de kantine. De manager heeft een gele peer gegeten, en de assistent heeft een groene peer gegeten. De manager heeft later die middag opnieuw een appel gegeten want ze wilde meer fruit eten.} \\ 
    \hline
    R\_MI\_MA & {\footnotesize De manager en de assistent haalden in de pauze een stuk fruit uit de kantine. De manager heeft een gele peer gegeten, en de assistent heeft een groene appel gegeten. De manager heeft later die middag opnieuw een appel gegeten want ze wilde meer fruit eten.}\\
    \hline
    R\_MA\_MI & {\footnotesize De manager en de assistent haalden in de pauze een stuk fruit uit de kantine. De manager heeft een gele appel gegeten, en de assistent heeft een groene peer gegeten. De manager heeft later die middag opnieuw een appel gegeten want ze wilde meer fruit eten.} \\ 
    \hline
    R\_MA\_MA & {\footnotesize De manager en de assistent haalden in de pauze een stuk fruit uit de kantine. De manager heeft een gele appel gegeten, en de assistent heeft een groene appel gegeten. De manager heeft later die middag opnieuw een appel gegeten want ze wilde meer fruit eten.} \\
    \hline
    P\_MI\_MI & {\footnotesize De assistent en de manager haalden in de pauze een stuk fruit uit de kantine. De assistent heeft een groene peer gegeten, en de manager heeft een gele peer gegeten. De manager heeft later die middag opnieuw een appel gegeten want ze wilde meer fruit eten.} \\
    \hline
    P\_MI\_MA & {\footnotesize De assistent en de manager haalden in de pauze een stuk fruit uit de kantine. De assistent heeft een groene appel gegeten, en de manager heeft een gele peer gegeten. De manager heeft later die middag opnieuw een appel gegeten want ze wilde meer fruit eten.}\\
    \hline
    P\_MA\_MI & {\footnotesize De assistent en de manager haalden in de pauze een stuk fruit uit de kantine. De assistent heeft een groene peer gegeten, en de manager heeft een gele appel gegeten. De manager heeft later die middag opnieuw een appel gegeten want ze wilde meer fruit eten.} \\ 
    \hline
    P\_MA\_MA & {\footnotesize De assistent en de manager haalden in de pauze een stuk fruit uit de kantine. De assistent heeft een groene appel gegeten, en de manager heeft een gele appel gegeten. De manager heeft later die middag opnieuw een appel gegeten want ze wilde meer fruit eten.} \\
    \hline
    \end{tabular}
    \caption{Sample sentences from `Again' Study 2.}
    \label{fig:dutch_discourse_2referents}
    \end{table*}

\end{document}